\documentclass[a4paper,twoside]{article}
\usepackage{natbib}
\usepackage{epsfig}
\usepackage{subcaption}
\usepackage{calc}
\usepackage{amssymb}
\usepackage{amstext}
\usepackage{amsmath}
\usepackage{amsthm}
\usepackage{multicol}
\usepackage{pslatex}
\usepackage{apalike}
\usepackage{SCITEPRESS}     

\usepackage{hyperref}
\usepackage[utf8]{inputenc} 

\usepackage{xcolor}
\usepackage{graphicx,import}
\usepackage{url}
\usepackage{algorithm}      
\usepackage{algpseudocode} 
\usepackage{booktabs}
\usepackage[section]{placeins}
\usepackage[switch]{lineno}
\usepackage[symbol]{footmisc}
\makeatletter

\def\medhline{%
	\noalign{\ifnum0=`}\fi\hrule \@height \medarrayrulewidth \futurelet
	\reserved@a\@xmedhline}
\def\@xmedhline{\ifx\reserved@a\medhline
	\vskip\doublerulesep
	\vskip-\medarrayrulewidth
	\fi
	\ifnum0=`{\fi}}
\makeatother



\newlength{\medarrayrulewidth}
\begin{document}
\title{Radar Artifact Labeling Framework (RALF): \\ Method for Plausible Radar Detections in Datasets}

\author{\authorname{
		Simon T. Isele,\sup{1,2,}\thanks{Authors contributed equally.}
		Marcel P. Schilling,\sup{1,2,a}
		Fabian E. Klein,\sup{1,a}\\
		Sascha Saralajew,\sup{1}  
		J. Marius Zoellner \sup{2,3}}
\affiliation{\sup{1}Dr. Ing. h.c. F. Porsche AG, Weissach, Germany.}
\affiliation{\sup{2}Karlsruhe Institute of Technology (KIT), Karlsruhe, Germany.}
\affiliation{\sup{3}FZI Research Center for Information Technology, Karlsruhe, Germany.}
\email{\{simon.isele1, fabian.klein\}@porsche.de, marcel.schilling@kit.edu, zoellner@fzi.de}
}


\abstract{Research on localization and perception for Autonomous Driving is mainly focused on camera and LiDAR datasets, rarely on radar data. Manually labeling sparse radar point clouds is challenging. For a dataset generation, we propose the cross sensor Radar Artifact Labeling Framework (RALF). Automatically generated labels for automotive radar data help to cure radar shortcomings like artifacts for the application of artificial intelligence. RALF provides plausibility labels for radar raw detections, distinguishing between artifacts and targets. The optical evaluation backbone consists of a generalized monocular depth image estimation of surround view cameras plus LiDAR scans. Modern car sensor sets of cameras and LiDAR allow to calibrate image-based relative depth information in overlapping sensing areas. K-Nearest Neighbors matching relates the optical perception point cloud with raw radar detections. In parallel, a temporal tracking evaluation part considers the radar detections' transient behavior. Based on the distance between matches, respecting both sensor and model uncertainties, we propose a plausibility rating of every radar detection. We validate the results by evaluating error metrics on semi-manually labeled ground truth dataset of $3.28\cdot10^6$ points. Besides generating plausible radar detections, the framework enables further labeled low-level radar signal datasets for applications of perception and Autonomous Driving learning tasks.}

\onecolumn \maketitle \normalsize \setcounter{footnote}{0} \vfill

\section{Introduction}
\label{sec:introduction}
\begin{figure*}
	\centering
	\includegraphics[width=\textwidth]{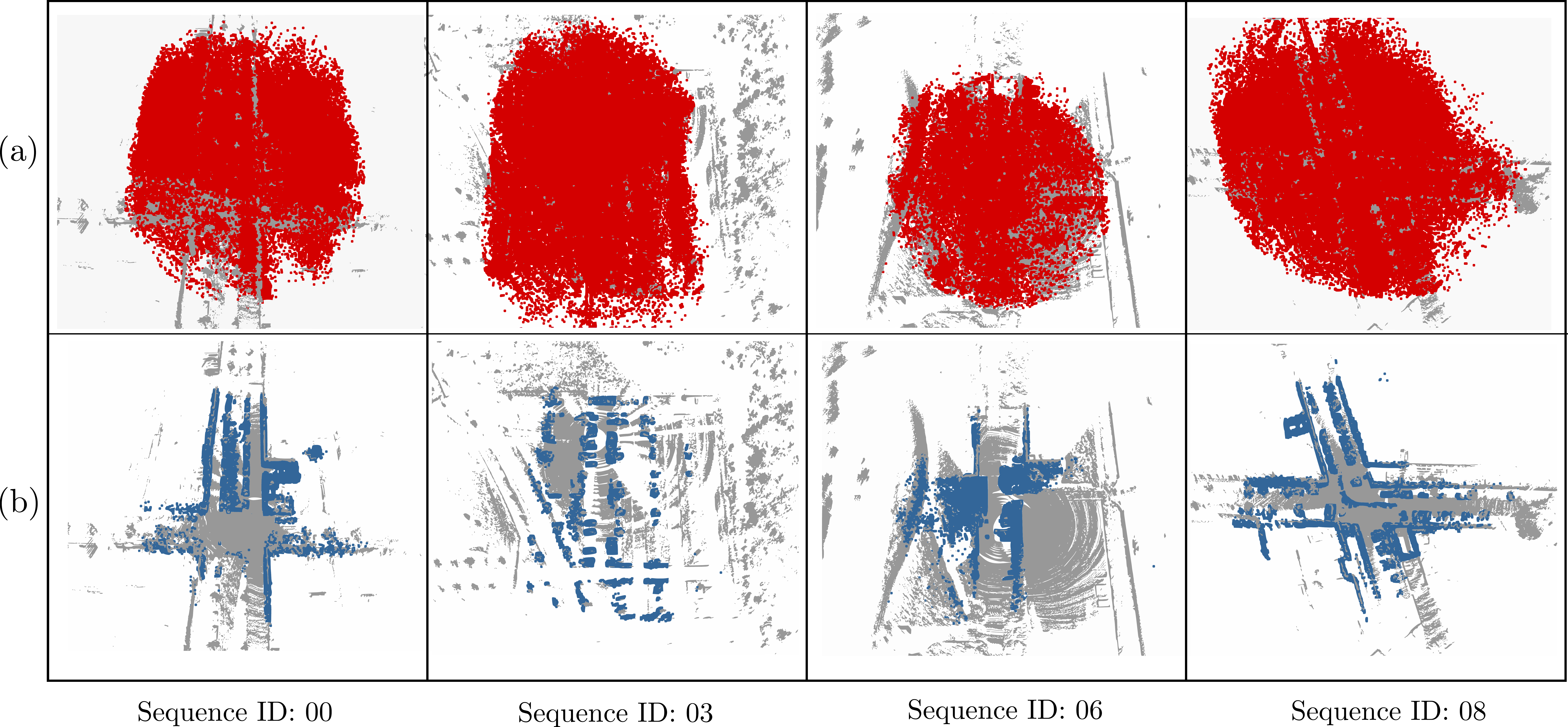}
	\caption{Scene comparison of four exemplary sequences with (a) raw radar detections (red) with underlying LiDAR (grey) and (b) manually corrected ground truth labels (blue) with plausible detections ($\hat{y} (\mathbf{p}_{i,t}) = 1 $).}
	\label{fig:scene_results}
\end{figure*} 
Environmental perception is a key challenge in the research field of Autonomous Driving (AD) and mobile robots. Therefore, we aim to boost the perception potential of radar sensors.
Radar sensors are simple to integrate and reliable also in adverse weather conditions~\citep{survey_autonomous_driving}. Post processing of reflected radar signals in the frequency domain, they provide 3D coordinates with additional information e.g. signal power or relative velocity. Such reflection points are called detections. But drawbacks such as sparsity~\citep{feng2019deep} or characteristic artifacts~\citep{holder2019modeling} call for discrimination of noise, clutter, and multi-path reflections from relevant detections.
Radar sensors are classically applied for Adaptive Cruise Control (ACC)~\citep{acc_radar} and state-of-the-art object detection~\citep{feng2019deep}. But to the authors best knowledge, radar raw signals are rarely used directly for AD or Advanced Driver Assistance Systems (ADAS).

Publicly available datasets comparable to KITTI~\citep{Geiger2013IJRR} or Waymo Open \citep{sun2019scalability} lack radar raw detections, and recently published datasets of nuScenes~\citep{nuscenes2019} or Astyx~\citep{astyx} are the only two available datasets containing both radar detections and objects respectively. However, transferability suffers from undisclosed preprocessing of radar signals e.\,g. \citep{nuscenes2019} or only front facing views~\citep{astyx}. 
Investigations for example on de-noising of radars by means of neural networks in supervised learning or other radar applications for perception in AD, currently require expensive and non-scaleable manually labeled datasets.

\begin{figure*}[t]
	\centering
	\def\svgwidth{0.99\textwidth}
	\fontsize{6.5pt}{11pt}
	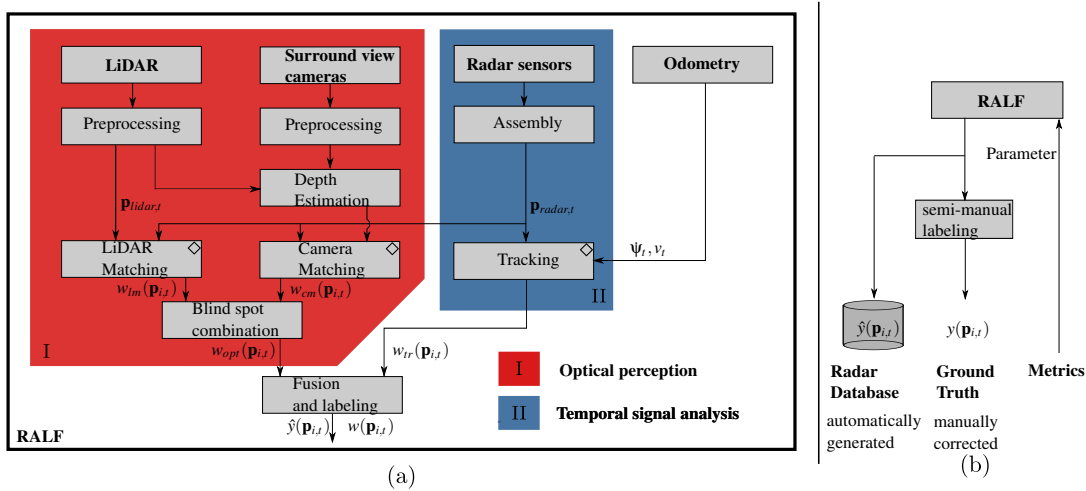
	\caption{Annotation pipeline RALF (a) with branches (I, II) and crucial components ($ \diamond $) as well as the overall method (b).}
	\label{fig:annotation_pipeline}
\end{figure*}

In contrast, we propose a generic method to automatically label radar detections based on on-board sensors denoted as Radar Sensor Artifact Labeling Framework (RALF). RALF enables a competitive, unbiased, and efficient data-enrichment pipeline as an automated process to generate consistent radar datasets including knowledge covering plausibility of radar raw detections. Inspired by Piewak et al.\,(2019), RALF applies the benefits of cross-modal sensors and is a composition of two parallel signal processing pipelines as illustrated in Figure~\ref{fig:annotation_pipeline}: Optical perception (I), namely camera and LiDAR as well as temporal signal analysis (II) of radar detections. Initially, false labeled predictions of RALF can be manually corrected, so one obtains hereby a ground truth dataset. This enables evaluation of RALF, optimization of its parameters, and finally unsupervised label predictions on radar raw detections. 

Our key \textbf{contributions} are the following: 

\begin{enumerate}
	\item An evaluation method to rate radar detections based on their existence plausibility using LiDAR and a monocular surround view camera system.
	\item A strategy to take the transient signal course into account with respect to the detection plausibility. 
	\item An automated annotation framework (RALF) to generate binary labels for each element of radar point cloud data describing its plausibility of existence, see Figure~\ref{fig:scene_results}.
	\item A simple semi-manual annotation procedure using predictions of RALF to evaluate the labeling results, optimize the annotation pipeline, and generate a reviewed radar dataset. 
\end{enumerate}

\section{Related work}
\noindent To deploy radar signals more easily in AD and ADAS applications, raw signal de-noising\footnote{We use the term de-noising to distinguish between plausible radar detections and artifacts, denoting no limitation only to noise in the classical sense. Our understanding of radar artifacts is based on the work of \cite{holder2019modeling}. To name an example, an artifact could be a mirror reflection, a target could be a typical object like a car, building or poles.} is compulsory. Radar de-noising can be done on different abstraction levels, at lowest on received reflections in the time or frequency domain (e.\,g. \cite{rock2019complex}).
At a higher signal processing level of detections in 3D space, point cloud operations offer rich opportunities. For instance, point cloud representations profit from the importance of LiDAR sensors and the availability of many famous public LiDAR datasets (\cite{sun2019scalability}; \cite{nuscenes2019}; \cite{Geiger2013IJRR}) in company with many powerful point cloud processing libraries such as pcl~\citep{rusu20113d} or open3D~\citep{zhou2018open3d}.
Transferred to sparse LiDAR point cloud applications, Charron et al. (2018) discussed shortcomings of standard filter methods (e.\,g. image filtering approaches to fail at sparse point cloud de-noising). 
DBSCAN \citep{ester1996density} is an adequate measure to cope with sparse noisy point clouds~\citep{daimler_dbscan}.
Point Cloud Libraries Libraries (e.\,g. Zhou et al., 2018; Rusu and Cousins, 2011) provide implementations of statistical outlier removal and radius outlier removal. Radius outlier removal is adapted considering proportionality between sparsity and distance~\citep{Charron2018DenoisingOL}, but the problem of filtering sparse detections in far range still remains unsolved.
To generate maps of the static environment with radar signals in a pose GraphSLAM~\citep{thrun2006graphslam}, \cite{radar_slam} applies RANSAC~\citep{ransac} and M-estimators~\citep{huber1964} to filter detections by their relative velocity information. Applying neural networks is an alternative strategy to filter out implausible points considering traffic scenes as a whole~\citep{Heinzler_2020}.

However, considering supervised deep learning approaches to be trained for filtering, ground truth labels are required. To the authors' best knowledge, there are no publicly available radar point cloud datasets explicitly enriched with raw point detection and related labels. Point-wise manual labeling is too time-consuming and therefore an automated labeling process is necessary. \cite{piewak2018boosting} developed an auto-labeling to create a LiDAR dataset. in this framework, the underlying idea is to make use of a state-of-the-art semantic segmentation model for a camera. With that, pixel-wise class information is associated to LiDAR detections by projection of the point cloud into the segmented image. The main correspondence problem for such a method are different Field of Views (FoVs), resulting in obstruction artifacts or differing aspect ratios. To compensate this, \cite{piewak2018boosting} suggested sensors to be mounted in closest possible proximity to avoid correspondence problems.
Recently, \cite{Behley_2019_ICCV}) published a semantically segmented LiDAR dataset, whose structural shell allows to transfer their workflow to other point cloud data.

\subsection{Method}
\noindent The proposed framework, visualized in Figure~\ref{fig:annotation_pipeline}, consists of an optical perception branch (I), namely camera and LiDAR, and temporal signal analysis (II). These branches are fused in the framework to output a consistent label of radar detections. RALF is implemented in the Robot Operating System~\citep{ros}.  
\subsection{Problem formulation and notation}
Inspired by other notations (\cite{fan2019pointrnn}; \cite{qi2017pointnet}), a point cloud at time $t$ is represented by spatial coordinates and corresponding feature tuples $\mathcal{P}_t = \left\lbrace (\mathbf{p}_{1,t},\mathbf{x}_{1,t}), ~\ldots, ~ (\mathbf{p}_{N_t,t},\mathbf{x}_{N_t,t}) \right\rbrace $, where $N_t$ denotes the total number of detections at time $t$. 
Spatial information is typically range $r$, azimuth angle $\varphi$, and elevation angle $\vartheta$. 
Additionally, radar detection specific information, e.\,g. Doppler velocity or signal power of the reflection, is contained in the feature vector $\mathbf{x}_{i,t} \in \mathbb{R}^{C}$ of point $\mathbf{p}_{i,t}$.  
The basic concept of the proposed annotation tool is to enrich each radar detection $\mathbf{p}_{i,t}$ with a corresponding feature attribute, namely plausibility $w(\mathbf{p}_{i,t}) \in \left[0,1 \right] $. The term plausibility describes the likelihood of a radar detection to represent an existing object ($y(\mathbf{p}_{i,t})=1$) or an artifact ($y(\mathbf{p}_{i,t})=0$).
\subsection{Annotation pipeline of RALF}
\label{sec:pipleline}
\begin{figure*}
	\centering
	\includegraphics[width=\textwidth]{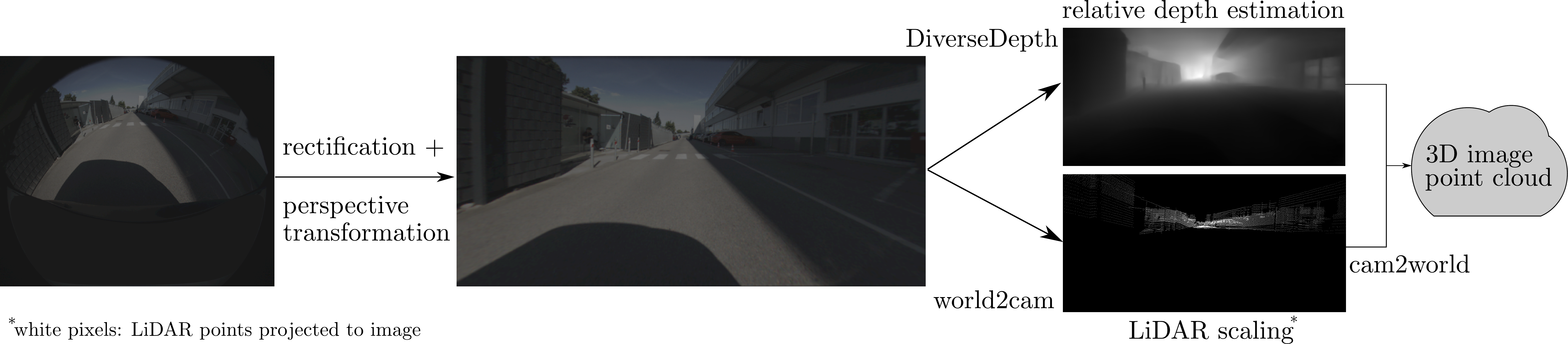}
	\caption{Camera matching pipeline.}
	\label{fig:cam_matching_pipeline}
\end{figure*}

\begin{figure}
	\centering
	\includegraphics[width=0.99\columnwidth]{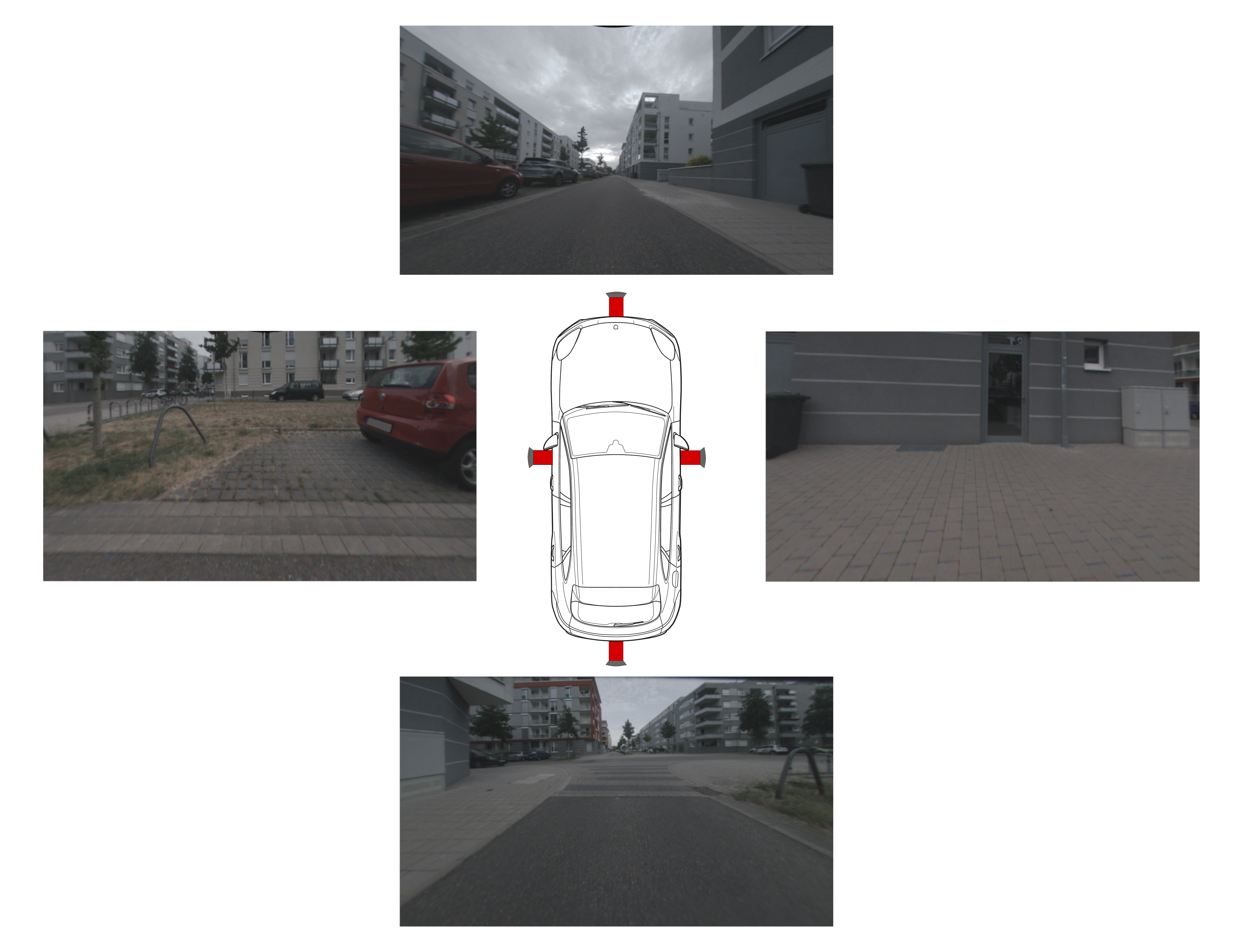}
	\caption{Preprocessed surround view camera in example scene.}
	\label{fig:scene}
\end{figure}

In the following, we describe the single modules that align a sensor signal with the reference system.
To enable comparison of multi-modal sensors, time synchronization~\citep{ros_synchro} and coordinate transformation~\citep{dillmann2013informationsverarbeitung} into a common reference coordinate system (see Figure~\ref{fig:sensor_setup}) are necessary.
\begin{algorithm}[t]
	\caption{LiDAR matching} 
	\label{alg:lidar_matching}
	\begin{algorithmic}
		\Require $\mathcal{P}_{\text{radar}, t}, \mathcal{P}_{\text{lidar}, t}  $ 
		\Ensure $w_{\text{lm}}(\mathbf{p}_{i,t}) $ 
		\For {$it=1,\ldots N_{\text{radar},t} $}
		\State $\mathbf{q} \gets$ \Call{k-NN}{$\mathcal{P}_{\text{lidar}, t},~ \mathbf{p}_{i,t},~K $}
		\State $d \gets 0$
		\For {$l=1,\ldots K $}
		\State $p_{x,l,t}, p_{y,l,t}, p_{z,l,t} \gets \mathcal{P}_{\text{lidar}, t}.\text{get\_point}(\mathbf{q} \left[ l\right] ) $ 
		\State $r_{l,t}, \varphi_{l,t}, \vartheta_{l,t} \gets \mathcal{P}_{\text{lidar}, t}.\text{get\_features}(\mathbf{q} \left[ l\right] ) $ 
		\State $\sigma_{d,i,l} \gets $
		\Call{model}{$r_{i,t}, \varphi_{i,t}, \vartheta_{i,t}, r_{l,t}, \varphi_{l,t}, \vartheta_{l,t} $}
		\State $d \gets d +\sqrt{\frac{\Delta p_{x,t}^2+\Delta p_{y,t}^2+\Delta p_{z,t}^2}{\sigma_{d,i,l}^2+ \epsilon}}$ 
		\EndFor	
		\State $w_{\text{lm}}(\mathbf{p}_{i,t}) \gets \exp(- \beta_{\text{lm}} ~ \frac{d}{K})$ 
		\EndFor
	\end{algorithmic} 
\end{algorithm} 
\paragraph{LiDAR matching}
\label{sec:lidar_matching}
This module aligns radar detections with LiDAR reflections as described in Algorithm~\ref{alg:lidar_matching}. Based on a flexible distance measure, plausibility of radar detections is determined. The hypothesis of matching reliable radar detections with LiDAR signals in a single point in space does not hold in general. LiDAR signals are reflected on object shells, while radar waves might also penetrate objects. Thus, some assumptions and relaxations are necessary in Algorithm~\ref{alg:lidar_matching}. We assume for the assessment no negative weather impact on LiDAR signals and comparable reflection modalities. Furthermore, since radar floor detections are mostly implausible, we estimate the LiDAR point cloud ground plane parameters via RANSAC~\citep{ransac} and filter out corresponding radar points.
Applying a k-Nearest Neighbor (k-NN) clustering (\cite{zhou2018open3d}; \cite{rusu20113d}) in Algorithm~\ref{alg:lidar_matching}, each radar detection $\mathbf{p}_{j,t}$ of the radar point cloud $\mathcal{P}_{\text{radar}, t}$, is associated with its $K$ nearest neighbors of the LiDAR scan  $\mathcal{P}_{\text{lidar}, t}$. Notice that values of $K$ greater than one improve the robustness due to less sparsity in LiDAR scans. Measurement equations  
\begin{eqnarray}
\label{eq:h_x}
h_x \:= & r_i ~ \cos \vartheta_i ~ \cos \varphi_i +  ~^\text{v}x_{\text{radar}}    \nonumber \\
 & - (r_l ~ \cos \vartheta_l ~ \cos \varphi_l +  ~^\text{v}x_{\text{lidar}}),\\
\label{eq:h_y}
h_y \: = & r_i ~ \cos \vartheta_i ~ \sin \varphi_i +  ~^\text{v}y_{\text{radar}} \nonumber \\
  & - (r_l ~ \cos \vartheta_l ~ \sin \varphi_l +  ~^\text{v}y_{\text{lidar}}),  \\
\label{eq:h_z}
h_z  = & r_i ~ \sin \vartheta_i +  ~^\text{v}z_{\text{radar}} - (r_l ~ \sin \vartheta_l +  ~^\text{v}z_{\text{lidar}})  
\end{eqnarray} 
are introduced as components of $L^2$ norm $d$ in Cartesian coordinates. Radar ($r_{i} ,\varphi_{i}, \vartheta_{i}$) and LiDAR detections ($r_{l} ,\varphi_{l}, \vartheta_{l}$) are initially measured in the local sphere coordinate system. Constant translation offsets $( ^\text{v}x_{\text{radar}}, ~^\text{v}y_{\text{radar}}, ~^\text{v}z_{\text{radar}})$ and $( ^\text{v}x_{\text{lidar}}, ~^\text{v}y_{\text{lidar}}, ~^\text{v}z_{\text{lidar}})$ relate the local sensor origins to the vehicle coordinate system. Assuming independence between uncertainties of radar coordinate measurements ($\sigma_{r, \text{radar}},  \sigma_{\varphi, \text{radar}}, \sigma_{\varphi, \text{radar}}$) as well as Time-of-Flight LiDAR uncertainty ($\sigma_{r, \text{lidar}}$), error propagation in Cartesian space can be obtained by
\begin{eqnarray}
\sigma_{d,i,l}^2 & = &  \bigg(\frac{\partial d}{ \partial r_i} \bigg)^2 \sigma_{r, \text{radar}} ^2 + \bigg(\frac{\partial d}{ \partial \varphi_i} \bigg)^2 \sigma_{\varphi, \text{radar}}^2 \nonumber \\ 
&  & +  \bigg(\frac{\partial d}{ \partial \vartheta_i} \bigg)^2 \sigma_{\vartheta, \text{radar}}^2 + \bigg(\frac{\partial d}{ \partial r_l} \bigg)^2 \sigma_{r, \text{lidar}} ^2    
\label{eq:error_prop_d}
\end{eqnarray}
denoted as \texttt{MODEL} in Algorithm \ref{alg:lidar_matching}. Rescaling the mismatch in each coordinate dimension enables the required flexibility in the LiDAR matching module. To ensure $w_{\text{lm}}(\mathbf{p}_{j,t}) \in \left[0, 1 \right]$ and also increase resolution in small mismatches $d$, an exponential decay function with a tuning parameter $\beta_{\text{lm}} \in \mathbb{R}^{+}$ is applied subsequently.
\paragraph{Camera matching}
Holding for general mountings, we undistort the raw images and apply a perspective transformation to obtain a straight view, Figures~\ref{fig:cam_matching_pipeline} and \ref{fig:scene}. We derive the modified intrinsic camera matrix $\mathbf{A} \in \mathbb{R}^{3 \times 3}$ for the undistorted image~\citep{ocamcalib}.
To match 3D radar point clouds with camera perception, a dense optical representation is necessary.

Structure-from-Motion (SfM)~\citep{orbslam} on monocular images reconstructs sparsely. Reconstruction is often incomparable to radar, due to few salient features in poorly structured parking scenarios e.\,g. plain walls in close proximity. Moreover, initialization issues in low-speed situations naturally degrades SfM to be reliable for parking.

Hence, we apply a pre-trained version of DiverseDepth~\citep{yin2020diversedepth} on pre-processed images to obtain relative depth image estimations. Thanks to the diverse training set and the resulting generalization, it outperforms other estimators trained only for front cameras on datasets such as KITTI~\citep{Geiger2013IJRR}. Thus, it is applicable to generic views and cameras. 
To match the depth image estimations to a metric scale, LiDAR detections in the overlapping FoV are projected into each camera frame considering $\mathbf{A}$ and extrinsic camera parameters $\mathbf{B} \in \mathbb{R}^{4 \times 4}$ (world2cam). The projected LiDAR reflections serve as sparse sampling points from which the depth image pixels are metrically rescaled. Local scaling factors outperform single scaling factors from robust parameter estimation (\cite{ransac}; \cite{huber1964}) at metric rescaling.
Equidistant samples of depth image pixels are point-wisely calibrated with corresponding LiDAR points via KNN. Figure~\ref{fig:cam_matching_pipeline} shows how the calibrated depth is projected back to world coordinates by the (cam2world) function.
Afterwards, the association to radar detections analogously follows Algorithm~\ref{alg:lidar_matching}, but considers the uncertainty of the depth estimation and its propagation in Cartesian coordinates.

Though, being aware of camera failure modes, potential model failures requests for manual review of automated RALF results. Measuring inconsistencies between camera depth estimation and extrapolated LiDAR detections or LiDAR depth in overlapping FoVs, indicate potential failures.  

\begin{figure}
	\centering
	\includegraphics[width=0.8\columnwidth]{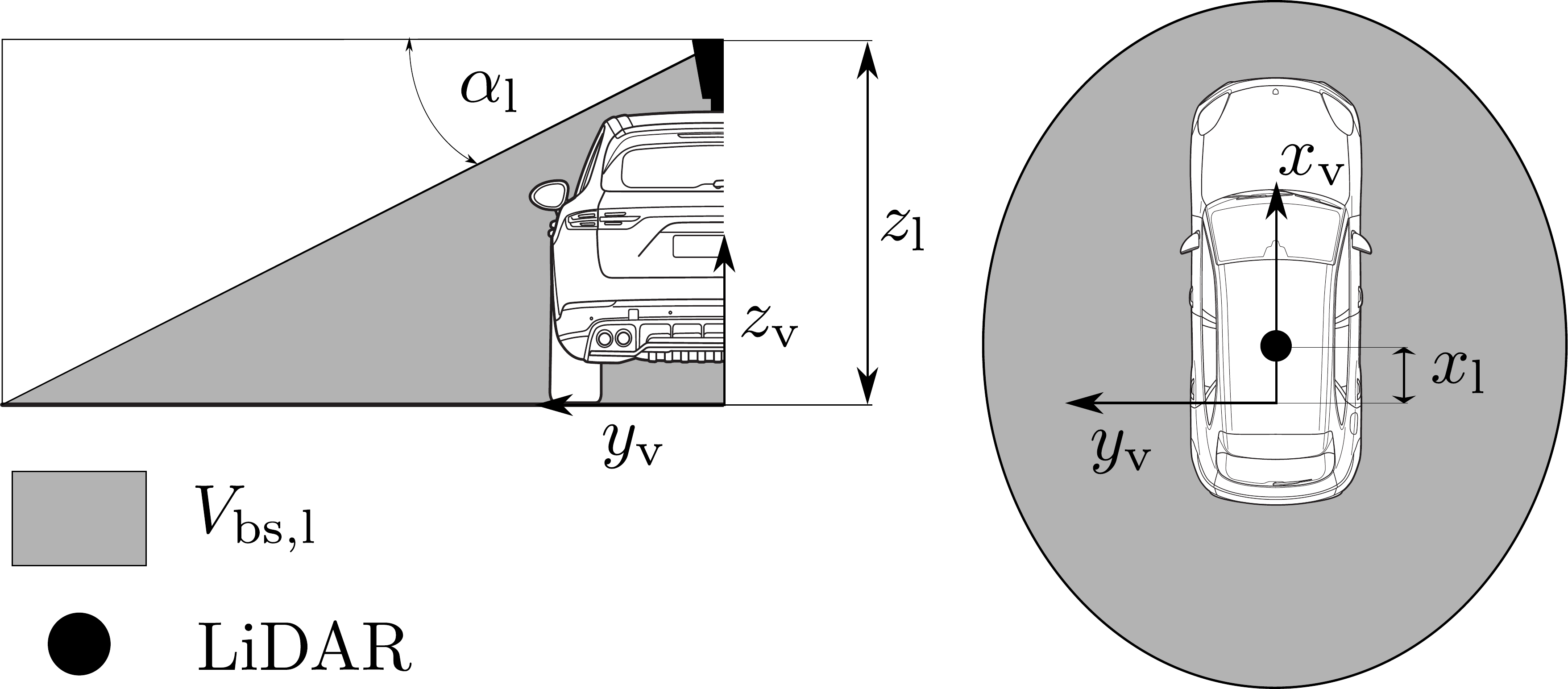}
	\caption{Blind spot LiDAR.}
	\label{fig:blindspot}
\end{figure}

\paragraph{Blind Spot Combination}
In the experimental setup, described in Section~\ref{sec:setup}, optical perception utilizing a single, centrally mounted LiDAR sensor lacks to cover the whole radar FoV as illustrated in Figure~\ref{fig:blindspot}. The set 
\begin{eqnarray}
V_{\text{bs,l}} &=& \bigg \lbrace \mathbf{p}_i = \left(p_{x,i},p_{y,i},p_{z,i}\right)^\top \in \mathbb{R}^3	  \mid   p_{z,i} \in \left[0, z_{\text{l}} \right]  \nonumber \\ 
&& ~~~ \wedge \sqrt{(p_{x,i}-x_{\text{l}})^2+p_{y,i}^2} \leq \frac{z_{\text{l}} -p_{z,i} }{\tan \alpha_{\text{l}}} \bigg\rbrace 
\label{eq:lidar_blindspot}
\end{eqnarray}
describes the LiDAR blindspot resulting from schematic mounting parameters ($y_{\text{l}}=0, z_{\text{l}}>0$) and opening angle $\alpha_{\text{l}}$ greater than zero.
Considering the different FoVs, 
\begin{equation}
w_{\text{opt}}(\mathbf{p}_{i,t}) = 
\begin{cases}
w_{\text{cm}}(\mathbf{p}_{i,t}) &  \mathbf{p}_{i,t} \in V_{\text{bs,l}} \\ 
w_{\text{lm}}(\mathbf{p}_{i,t})& \text{otherwise} \\
\end{cases}
\end{equation}
summarizes the plausibility of the optical perception branch (I).
Far range detection relies only on LiDAR sensing, while only cameras sense the nearfield. At overlapping intermediate sensing ranges, both rankings from camera and LiDAR scan are available instead of being mutually exclusive. Experiments yielded more accurate results for this region by preferring LiDAR over camera instead of a compromise of both sensor impressions.

\paragraph{Tracking}
Assuming Poisson noise~\citep{poisson} on radar detections, it is probable that real existing objects in space form hot spots over consecutive radar measurement frames, whereas clutter and noise is almost randomly distributed. Since labeling is not necessarily a real-time capable online process, one radar point cloud $\mathcal{P}_{\text{radar}, t_k}$ at $t_k$ forms the reference to evaluate spatial reoccurence of detections in radar scan sequences. Therefore, a batch of $n_{\text{b}} \in \mathbb{N}$ earlier and subsequent radar point clouds are buffered around the reference radar scan at time $t_k$. 
Considering low speed planar driving maneuvers, applying a kinematic single-track model based on wheel odometry is valid~\citep{werling}. Based on the measured yaw rate $\dot{\psi}$, the longitudinal vehicle velocity $v$, and the known time difference $\Delta t$ between radar scan $k+1$ and $k$, the vehicle state is approximated by 
\begin{eqnarray}
\label{eq:single_track}
\big(
x_{\text{v}} , y_{\text{v}}, z_{\text{v}} ,\psi
\big)_{k+1}^\top = 
\big(
x_{\text{v}} , y_{\text{v}},  z_{\text{v}} ,\psi
\big)_{k}^\top \nonumber \\
+ \Delta t \cdot
\big(
v~\cos \psi ,v~\sin \psi , 0, \dot{\psi}
\big)_k^\top.
\end{eqnarray}

Considering Equation~\eqref{eq:single_track} and rotation matrix $ R_{z,\psi}$, containing yaw angle $\psi$, allows ego-motion compensation for each point $i$ of the buffered radar point clouds $\mathcal{P}_{\text{radar}, t_{k+j}}$ for $j \in \lbrace -n_{\text{b}}, \ldots , -1,~ 1, \ldots , n_{\text{b}} \rbrace$ to the reference cloud $\mathcal{P}_{\text{radar}, t_k}$. 
Each point
\begin{eqnarray}
\label{eq:pose_trafo}
\mathbf{\tilde{p}}_{i,t_{k+j}} \:= \: R_{z,\psi_k }^{-1}\Big (\big( 
x_{\text{v}} , y_{\text{v}},  z_{\text{v}}
\big)_{k+j}^\top -
\big(x_{\text{v}} , y_{\text{v}},  z_{\text{v}}
\big)_{k}^\top \Big) \nonumber \\ 
+ \; \; R_{z,\psi_{k+j}-\psi_{k}}~ \mathbf{p}_{i,t_{k+j}} 
\end{eqnarray}
represents the spatial representation of a consecutive radar scan after ego-motion compensation to the reference radar scan at time step $t_k$. Enabling temporal tracking and consistency checks on the scans, $n_{\text{b}}$ should be chosen regarding the sensing cycle time. Assuming to analyze mainly static objects, Equation~\eqref{eq:pose_trafo} is valid. To describe dynamic objects, Equation~\eqref{eq:pose_trafo} has to be extended considering Doppler velocity and local spherical coordinates of each detection. By taking error propagation in the resulting measurement equations into account, different uncertainty dimensions are applicable. We apply spatial uncertainty as in Equation~\eqref{eq:error_prop_d}. The analysis of $n_{\text{b}}$ scans result in a batch of distance measures $d_j$. Simple averaging fails due to corner-case situations in which potentially promising detections remain undetected in short sequences. Hence, sorting $d_j$ in ascending order and summing the sorted distances, weighted by a decreasing coefficient with increasing position, yields promising results.

\paragraph{Fusion and final labeling}
\begin{figure}
	\centering
	\includegraphics[width=.36\columnwidth]{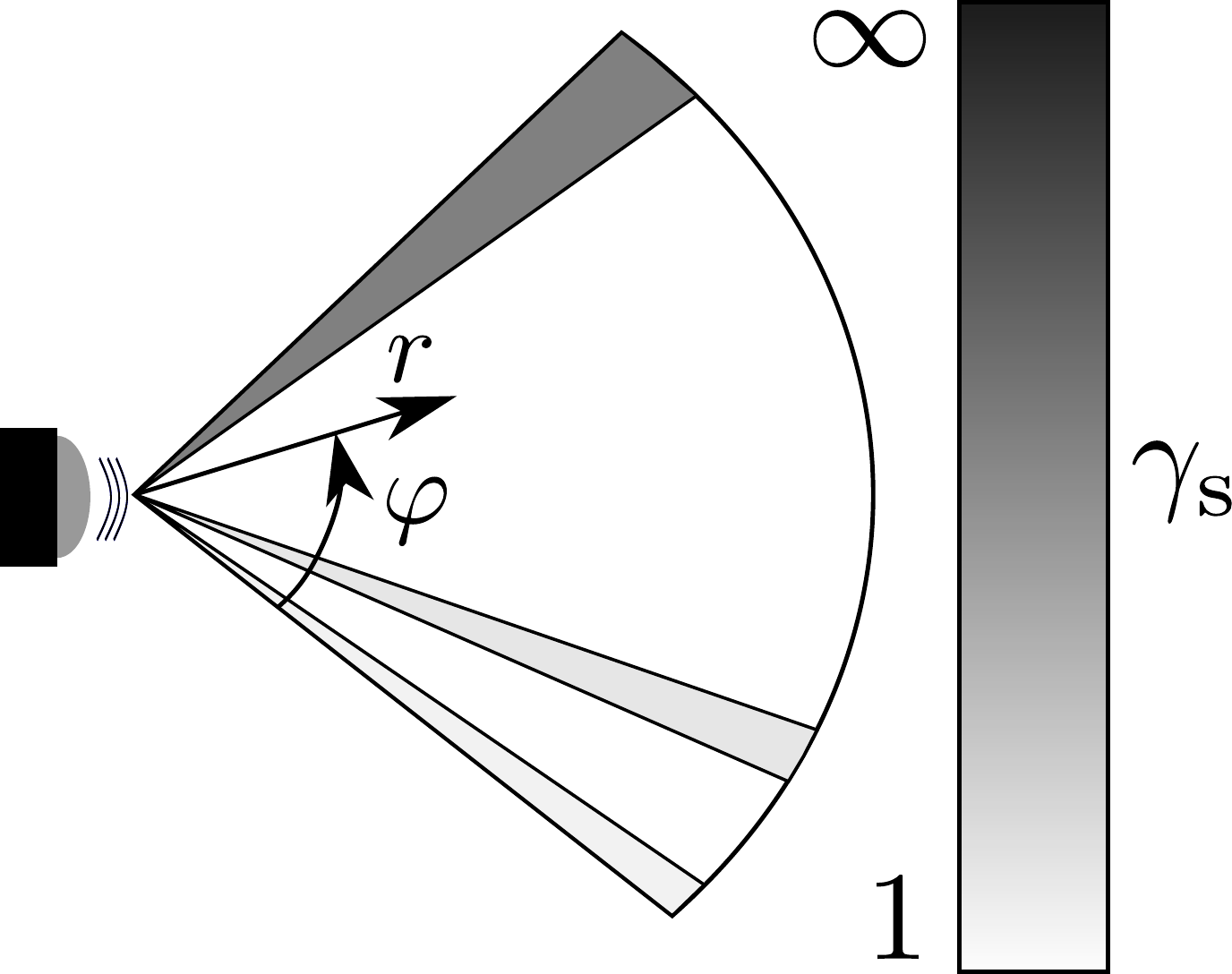}
	\caption{Reliabilty scaling.}
	\label{fig:gamma}
\end{figure}
The outputs of optical perception and tracking are combined with a setup-specific sensor a-priori information $\gamma_{\text{s}}(\varphi) \in \left[1, \infty \right)$ is combined. 
Since radar sensors are often covered behind bumpers, inhomogenous measurement accuracies $\gamma_{\text{s}}(\varphi_{i,t})$ arise over the azimuth range $\varphi$, see Figure~\ref{fig:gamma}. The a-priori known sensor specifics are modeled by the denominator in Equation \eqref{eq:priorization}. 

The tuning parameter $\alpha \in \left[0,1 \right]$ prioritizes between the tracking and optical perception module, formalized as first term $w(\mathbf{p}_{i,t})$ of the Heaviside function $H: \mathbb{R} \rightarrow \{0,1\}$ argument in Equation~\eqref{eq:priorization}. The final binary labels to discriminate artifacts ($y=0$) from promising detections ($y=1$) are obtained by
\begin{eqnarray}
\label{eq:priorization}
\hat{y}(\mathbf{p}_{i,t}) \;=& H\big(w(\mathbf{p}_{i,t})-w_{0}\big) \nonumber \\ =&
H\Bigg(\frac{\alpha~ w_{\text{opt}}(\mathbf{p}_{i,t}) + (1-\alpha) ~w_{\text{tr}}(\mathbf{p}_{i,t})}{\gamma_{\text{s}}(\varphi_{i,t})} -w_{0}\Bigg)  ,
\end{eqnarray} 
where $w_{0} \in \left[0,1\right]$ ia a threshold on the prioritized optical perception and tracking results. 

\subsection{Use-case specific labeling policy}
\label{sec:param_selection}
\begin{figure}
	\centering
	\includegraphics[width=.8\columnwidth]{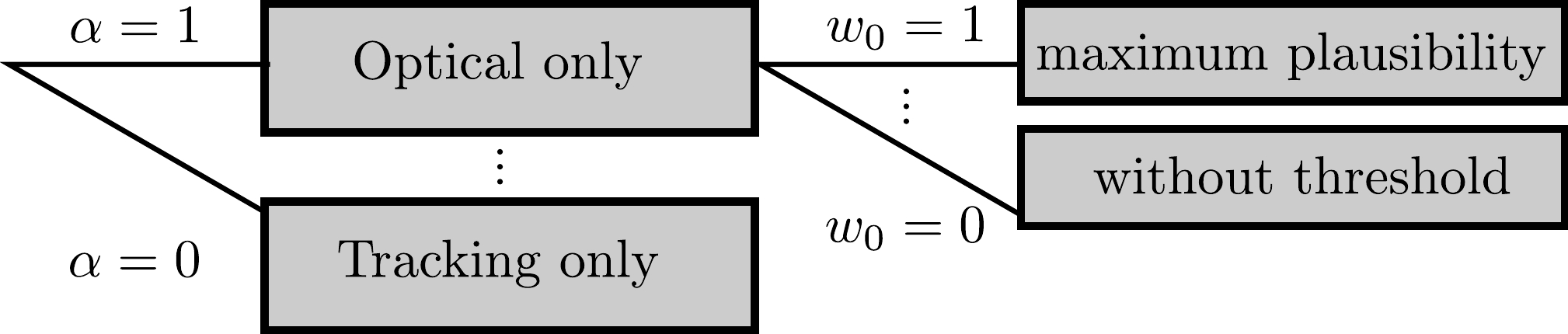}
	\caption{Parameter selection in RALF for branch weights $\alpha$ and plausibility threshold $w_0$.}
	\label{fig:param_selection}
\end{figure}

Motives for labeling a dataset might vary with the desired application purpose, along with conflicting parameter selection for some use cases. Our framework parameters allow to tune the automated labeling. High $\alpha=1$ suppresses radar detections without LiDAR detections or camera detections in their neighborhood, while low $\alpha=0$ emphasizes temporal tracking over the visual alignment, see Figure~\ref{fig:param_selection}.  
                          
Low $\alpha=0$ settings include plausible detections to occur behind LiDAR reflections, e.\,g. reflecting from inside a building. But, plausible radar detections are required to be locally consistent over several scans. On the upper bound $\alpha=1$, the temporal tracking consistency constraint vanishes its influence on the plausible detections, resulting in an optical filtering. For instance,  plausibility is rated high around LiDAR and camera perception.
Examples for the relevance of temporal tracking might be the localization on radar. To recognize a known passage, the scene signature and temporal sequence of radar scans might be much more important than a de-noised representation of a scene. The other extreme might be the use-case of semantic segmentation on radar point clouds where one is interested in the nearest and shell describing radar reflections, omitting reflections from the inside of objects.
Parameter $w_0$ acts as threshold margin on the plausibility. 

RALF parameters $\alpha, \beta_{\text{lm}}, \beta_{\text{cm}}, \beta_{\text{tr}}, n_{\text{b}}$ and $K$ can be tuned by manual inspection and according to the desired use-case. The error metrics in Equation~\eqref{eq:metricstart}-\eqref{eq:metricsend}, introduced in the Appendix, help to finetune the parametrization as visualized in Figure~\ref{fig:annotation_pipeline}(b).

Optimizing RALF subsequentially leads to more accurate predictions and decaying manual label corrections. 
However, proper initial parameters and having a manageable parameter space is essential. 
After this fine-tuning step, no further manual parameter inspection of RALF is necessary. The automatically predicted radar labels can be directly used to annotate the dataset.

We tune the desired performance to achieve an overestimating function. Manual correction benefits of coarser estimates that can be tailored to ground truth labels, whereas extension of bounds requires severe interference with clutter classifications. Hence, in Table~\ref{tab:data_set}, we aim for high Recall values while allowing lower Accuracy.
\subsection{Error evaluation}
\label{sec:error}
An error measure expressing the quality of the automated labeling is essential in two aspects, namely to check if the annotation pipeline is appropriate in general and to optimize its parameters. Since this paper proposes a method to generate plausibility of radar detections, it is challenging to describe a general measure that evaluates the results. 
Without ground truth labels, only indirect metrics are possible. For instance distinctiveness, expressed as difference between means of weights per class in combination with balance of class members. However, several cases can be constructed in which this indirect metric misleads. Therefore, we semi-manually labeled a set of $\text{M}=11$ different scans assisted by RALF to correctly evaluate the results, see Section~\ref{sec:semi_man_label} and Table~\ref{tab:data_set}.
\begin{figure}[t]
	\centering
	\includegraphics[width=.8\columnwidth]{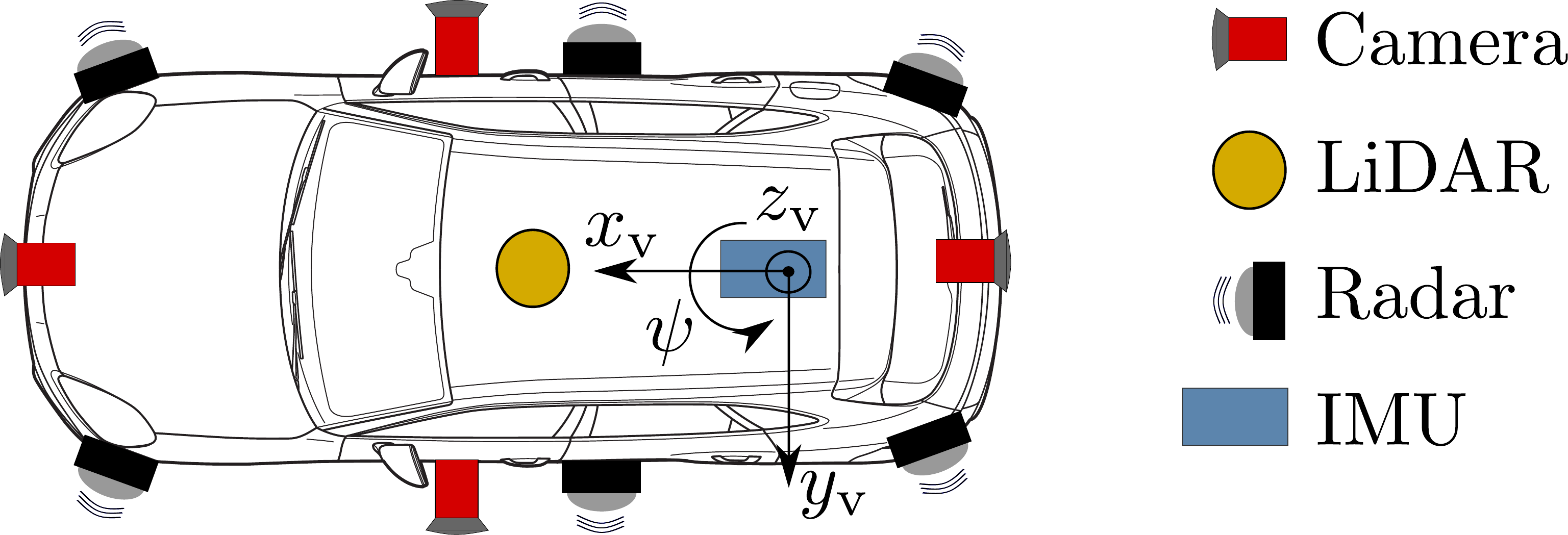}
	\caption{Sensor setup.}
	\label{fig:sensor_setup}
\end{figure}
\begin{table}[t]
	\caption{Sensor setup details.}
	\label{sample-table}
	\centering
	\begin{tabular}{lc}
		\toprule[1.5pt]
		Name & Details \\ 
		\toprule[0.8pt]
		Camera & 4 x Monocular surround view camera \\
		& (series equipment) \\ 
		LiDAR & Rotating Time-of-Flight LiDAR \\ & (centrally roof-mounted, 40 channel) \\ 
		Radar & 77~GHz FMCW Radar \\ & (FoV: $160$ degree hor., $\pm 10$ degree hor.)  \\ 
		\bottomrule[1.5pt]
	\end{tabular}
\end{table}

\section{Experiments}
In the first section, we describe the hardware sensor setup for the real world tests. After that, we discuss how the labeling policy depends on the use-case, and finally evaluate the results of the real world test.

\subsection{Sensor setup and experimental design}
\label{sec:setup}
The vehicle test setup is depicted in Figure~\ref{fig:sensor_setup} and sensor set details are found in Table~\ref{sample-table}. 
We evaluate the radar perception for a radar-mapping parking functionality. Eleven reference test tracks (e.\,g. urban area, small village, parking lot and industrial park) were considered and are depicted in the Appendix. To ensure a balanced, heterogeneous dataset, they contain parking cars, vegetation, building structures as well as a combination of car parks (open air and roofed). Vehicle velocity $v$ below $10$kph during the perception of the test track environment ensures large overlapping areas in consecutive radar scans.   

\subsection{Semi-manual labeling using predictions of RALF}
\label{sec:semi_man_label}
We use the predictions of RALF as a first labeling guess for which humans are responsible to correct false positives and false negatives. To ensure accurate corrections of RALF predictions, we visualize both radar and LiDAR clouds in a common reference frame and consider all parallel cameras to achieve ground truth data semi-manually.
\begin{table}[h!]
	\caption{Test set class balance over $\text{N}=2704$ radar scans.}
	\label{tab:class_balance}
	\centering
	\begin{tabular}{cccc}
		\toprule[1.5pt]
		$\Sigma \text{N}$  & $\frac{\varnothing \text{N}(y=1)}{\varnothing \text{N}}$ & $\frac{\varnothing \text{N}(y=0)}{\varnothing \text{N}}$  \\ 
		\toprule[0.8pt] 
		3 288 803 & 21.55~\%  & 78.45~\% \\ 
		
		\bottomrule[1.5pt]
	\end{tabular} 
\end{table}

\noindent We base the evaluation on a dataset of eleven independent test tracks for which we compare the RALF labels versus the manually corrected results. Containing two classes, the overall class balance of the dataset is $78.45\%$ clutter (2 580 066 radar points) against $21.55\%$ plausible detections (708 737 radar points).  Details can be found in Table~\ref{tab:data_set}. Additional imagery info and dataset statistics are found in the appendix.

Please note, following our manual correction policy, radar reflections of buildings are reduced to their facade reflections. Intra-building reflections are re-labeled as clutter although the detections might correctly result from inner structures. This results in heavily distorted average IoU values of plausible detections ranging from $36.7\%$ to $ 60.9\%$, Table~\ref{tab:data_set}. This assumption is essential for re-labeling and manual evaluation of plausible detections. Intra-structural reflections are hard to rate in terms of plausibility. Besides, in a transportation application, the sensory hull detection of objects needs to be reliable.

\begin{table*}[t]
	\caption{Test dataset of $\text{M}=11$ sequences.}
	\label{tab:data_set}
	\centering\small
	\begin{tabular}{ccccccccc}
		\toprule[1.5pt]
		
		ID 		& $\varnothing$N  & $\varnothing$Acc & $\varnothing$Precision  & $\varnothing$Recall & F1 plausible & $\varnothing$IoU & IoU plausible & IoU artifact  \\ 
		\toprule[1.5pt]
		& $\Sigma$~2704 & 0.873 &  0.826& 0.779& 0.675 & 0.682 & 0.510 & 0.854  \\
		\toprule[0.8pt]
		00 		& 245 & 0.848  &  0.833& 0.793& 0.722& 0.687& 0.565 & 0.810   \\ 
		01 		& 290 & 0.883  &  0.796& 0.781& 0.646& 0.673& 0.477 & 0.869  \\  
		02 		& 101 & 0.878  &  0.839& 0.822& 0.740& 0.720& 0.588 & 0.852   \\  
		03 		& 400 & 0.885  &  \textbf{0.706}& 0.761& 0.622& 0.662& 0.451 & 0.873   \\ 
		04 		& 334 &\textbf{0.927}  &  0.863& \textbf{0.859}& \textbf{0.765}& \textbf{0.768}& \textbf{0.620} & \textbf{0.917} \\  
		05 		& 163 & 0.910  & \textbf{ 0.869}& 0.845& 0.757& 0.752& 0.609 & 0.895  \\ 
		06 		& 170 & \textbf{0.776}  &  0.771& \textbf{0.678}& \textbf{0.537}& \textbf{0.554}& \textbf{0.367} & \textbf{0.742}  \\  
		07 		& 422 & 0.860  &  0.808& 0.739& 0.616& 0.644& 0.445 & 0.824  \\  
		08 		& 265 & 0.850  &  0.759& 0.771& 0.622& 0.640& 0.451 & 0.829  \\  
		09 		&  82 & 0.845  &  0.813& 0.776& 0.684& 0.667& 0.520 & 0.814  \\  
		10 		& 232 & 0.874  &  0.844& 0.798& 0.715& 0.703& 0.556 & 0.850  \\  
		
		\bottomrule[1.5pt]
	\end{tabular} 
\end{table*}
Using the data format of SemanticKitti dataset for LiDAR point clouds~\citep{Behley_2019_ICCV}, the evaluation of reliable radar detections orientates on the following clusters: $\texttt{human}$, $\texttt{vehicle}$, $\texttt{construction}$, $\texttt{vegetation}$, $\texttt{poles}$. Our proposed consolidation of original SemanticKITTI classes to a reduced number of clusters is found in the Appendix, see Table~\ref{tab:consolidation}. Especially the $\texttt{vegetation}$ class imposes labeling consistency. E.\,g. grass surfaces can be treated as relevant if the discrimination of insignificant reflections from ground seems possible. On the other hand, grass and other vegetation are source of cluttered, temporally and often also spatially unpredictable reflections. 
   
Different labeling philosophies impose the necessity of a consistent labeling policy. In the discussed dataset of this work, we emphasize on grass surfaces as plausible radar reflections in order to discriminate green space from road surface. Stuctural reflections are labeled based on facade reflections, while intra-vehicle detections are permitted as relevant.  Please note, road surface reflections are labeled as clutter.
Table~\ref{tab:class_balance} shows the class distribution of the labeled reference test of $\text{M}=11$ sequences with in average $\varnothing \text{N}=1216$ radar detections for evaluation in Section~\ref{sec:results}.
Inspecting Table~\ref{tab:class_balance}, please note, that the dataset has imbalanced classes, so a detection is more likely to be an artifact.

\subsection{Results}
The following section discusses the results on real world data and includes an evaluation.

\paragraph{Monocular Depth Estimation}
\label{sec:results}

\begin{figure}[h!]
	\centering
	\includegraphics[width=0.99\columnwidth]{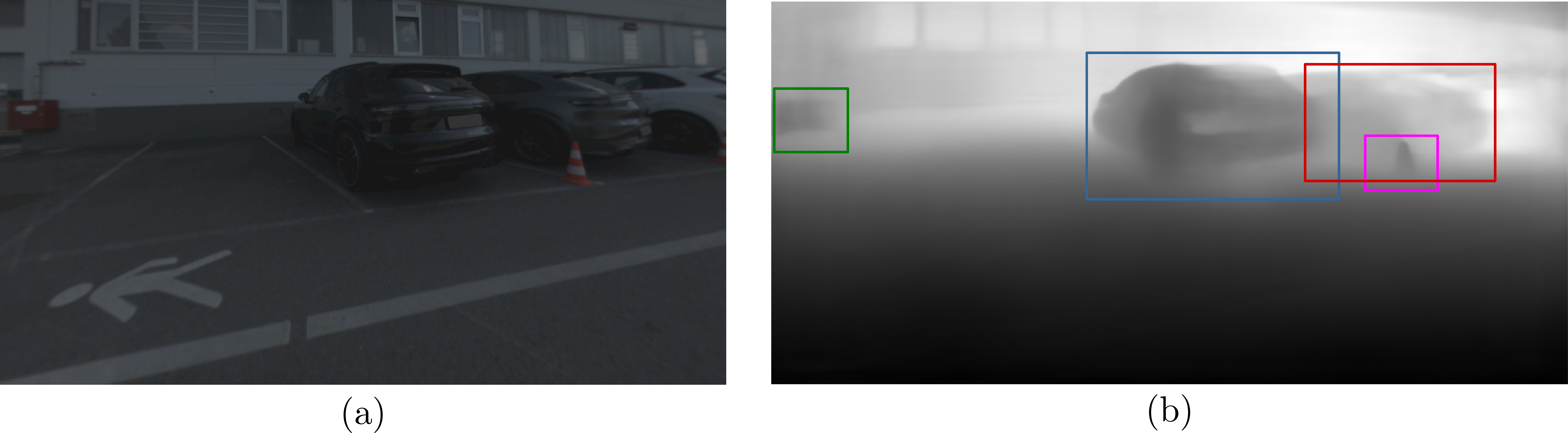}
	\caption{Depth estimation (b) with highlighted objects visible in the input image (a); an increasing depth is encoded by an increasing brightness in (b).}
	\label{fig:cam_result_left}
\end{figure}

\begin{figure}[h!]
	\centering
	\includegraphics[width=0.99\columnwidth]{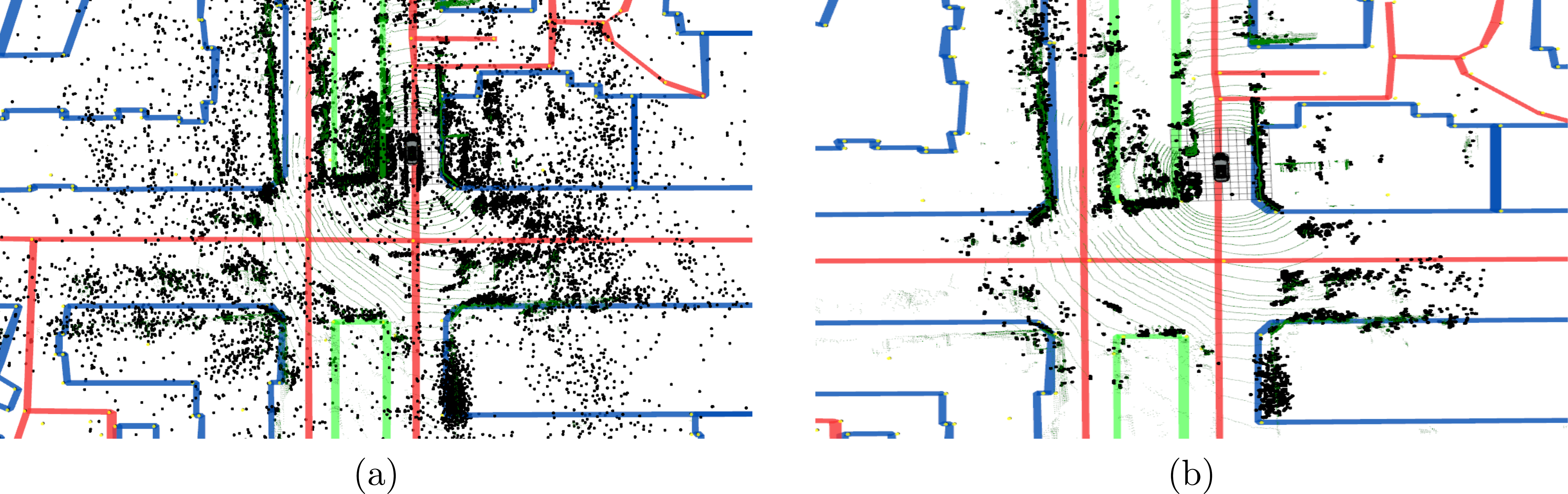}
	\caption{Birds-eye-view on example a scene with colored OpenStreetMap data~\citep{OpenStreetMap}. Comparison accumulated radar scans (black points): (a) all detections (b) plausible RALF detections ($\hat{y} (\mathbf{p}_{i,t}) = 1 $) without relabeling.}
	\label{fig:overview}
\end{figure}

Figure \ref{fig:cam_result_left} illustrates that the pre-trained depth estimation network provides fair results on pre-processed surround view cameras. 
Key contribution to achieve reasonable results on fisheye images without retraining are a perspective transformation and undistortion. However, in very similar scenes as shown in Figure~\ref{fig:cam_result_left}, it is challenging to estimate the true relative depth. 
By using local LiDAR scales as we propose, this issue can be solved elegantly and thus an overlap of LiDAR and camera FoV is a helpful benefit. Moreover, the results in Figure~\ref{fig:cam_matching_pipeline} and Figure~\ref{fig:cam_result_left} show that the depth estimation network generalizes to other views and scenes.
\paragraph{Qualitative Comparison Raw vs. Labeled Data}
Figure~\ref{fig:overview} illustrates an example scene. The results of RALF in this scene compared to unlabeled raw data are shown in Figures~\ref{fig:scene} and \ref{fig:overview}. Scene understanding is considerably facilitated.
\paragraph{Quantitative Evaluation}
\begin{table}
	\caption{Confusion matrix on dataset with $\Sigma\text{N}$ detections.}
	\label{tab:conf_mat}
	\centering
	\begin{tabular}{lcc}
		\toprule[1.5pt]
		& $y = 1$  & $y = 0$  \\ 
		\toprule[0.8pt] 
		$\hat{y} = 1$  & 2 432 440 & 268 869\\ 
		$\hat{y} = 0 $ & 157 076 & 430 418\\ 
		\bottomrule[1.5pt]
	\end{tabular} 
\end{table}
We use the prefaced manually labeled test set to evaluate the proposed pipeline. The confusion matrix of the dataset are found in Table~\ref{tab:conf_mat}.
By achieving a mean error of $L = 1-\text{Accuracy} = 12.95~\%$ on the dataset, we demonstrate the capability of the proposed pipeline to generate meaningful labels on real world test tracks. Please note the beneficial property of overestimation.
Comparing an average Recall of $77.9~\%$ to an average Precision $82.6~\%$ in Table~\ref{tab:data_set}, there is no preferred error in the labeling pipeline of RALF. Since RALF can be parameterized, increasing Recall and decreasing Precision or vice versa is possible by tuning the introduced parameters $w_0$ and $\alpha$.
Inspecting the differences of the performance per sequence, sequence~06 is exemplary for the lower performance, while sequence~04 performs best. Interestingly, these two sequences overlap partly. Sequence~06, including an exit of a narrow garage, poses difficulties in the close surrounding of the car, which explains the performance decrease.  
\paragraph{Robustness}
Errors can be provoked by camera exceptions (lens flare, darkness, etc.) and assumption violations. Near-field reconstruction results suffer in cases when ground and floor-standing objects in low height can not distinguished accurately, yielding vague near-field labels. Furthermore, in non-planar environments containing e.\,g. ascents, the planar LiDAR floor extraction misleads. This causes RALF to mislabel radar floor detections. Moreover, the tracking module suffers at violated kinematic single-track model assumptions.
\section{Conclusion}
We propose RALF, a method to rate radar detections concerning their plausibility by using optical perception and analyzing transient radar signal course. By a combination of LiDAR, surround view cameras, and DiverseDepth~\citep{yin2020diversedepth}, we generate a 360 degree perception in near- and far-field. DiverseDepth~\citep{yin2020diversedepth} yields a dense depth estimation, outperforming SfM approaches. Monitored via LiDAR, failure modes can be detected. Since considering model and sensor uncertainties respectively, a flexible comparison using different sensors is possible. From the optical perception branch, radar detections can be enriched by LiDAR or camera information as a side effect. Such a feature is useful for developing applications using annotated radar datasets. To evaluate RALF and fine-tune its parameters, RALF predictions can be semi-manually corrected to ground truth labels. Recorded vehicle measurements on real-world test tracks yield an average Accuracy of 87.3$\%$ at average Precision of 82.6$\%$ of the proposed labeling method, though satisfying de-noising capabilities. The evaluation reveals positive effects of an overestimating labeling performance. Time and effort for labeling are reduced significantly. As side notice, the labeling policy is coupled with the desired use-case and evaluation metrics which may differentiate.
We plan to extend the work on the framework towards semantic labeling.

\label{sec:conclusion}

\bibliographystyle{apalike}
{\small \bibliography{vehits21_ralf}}

\section*{Appendix}

\subsection*{Class Consolidation}
The authors of SemanticKITTI~\citep{Behley_2019_ICCV} introduce a class structure in their work. To transfer this approach to radar detections, we propose a consolidation of classes as found in Table~\ref{tab:consolidation}.
\begin{table*}
	\caption{Proposed clustering of SemanticKITTI classes~\citep{Behley_2019_ICCV}  to determine radar artifacts.}
	\centering
	\label{tab:consolidation}
	\begin{tabular}{lc}
		\toprule[1.5pt]
		\textbf{Cluster} & \textbf{SemanticKITTI Classes} \\ 
		\midrule[1pt]
		\textbf{Vehicle}  & \texttt{car, bicycle, motorcycle, truck, other-vehicle, bus}  \\ 
		\textbf{Human} & \texttt{person, bicyclist, motorcyclist}   \\ 
		\textbf{Contruction}  & \texttt{building, fence}   \\ 
		\textbf{Vegetation}  & \texttt{vegetation, trunk, terrain}   \\ 
		\textbf{Poles}  & \texttt{pole, traffic\_sign, traffic\_light}   \\ 
		\textbf{Artifacts}  & \texttt{sky, road, parking, sidewalk, other-ground}  \\ 
		\bottomrule[1.5pt]
	\end{tabular} 
\end{table*}

\subsection*{Metrics}
The applied metrics in Table~\ref{tab:data_set} are formulated based on the state-of-the art binary classification metrics True Positive (TP), False Positives (FP), True Negatives (TN), and False Negatives (FN).
\begin{equation}
\text{Accuracy} = \frac{TP + TN}{TP + TN + FP + FN}
\label{eq:metricstart}
\end{equation}
\begin{equation}
\text{Precision} = \frac{TP}{TP + FP}
\end{equation}
\begin{equation}
\text{Recall} = \frac{TP}{TP + FN}
\end{equation}
\begin{equation}
\text{F1} = \frac{2 \cdot TP}{2 \cdot TP + FP + FN}
\end{equation}
The metric mean Intersection-over-Union ($\varnothing$IoU) is based on the mean Jaccard  Index~\citep{IoU_2015} which is normalized over the classes C. The metric IoU expresses the labeling performance class-wise. 
\begin{equation}
\varnothing \text{IoU} = \frac{1}{C} ~ \sum\limits_{c=1}^C ~ \frac{TP_{c}}{TP_{c} + FP_{c} + FN_{c}}
\label{eq:metricsend}
\end{equation}

\subsection*{Sequence description}
The sampled sequences are shortly introduced for visual inspection and scene unterstanding.

\begin{figure}[h!]
	\centering
	\includegraphics[width=0.83\columnwidth]{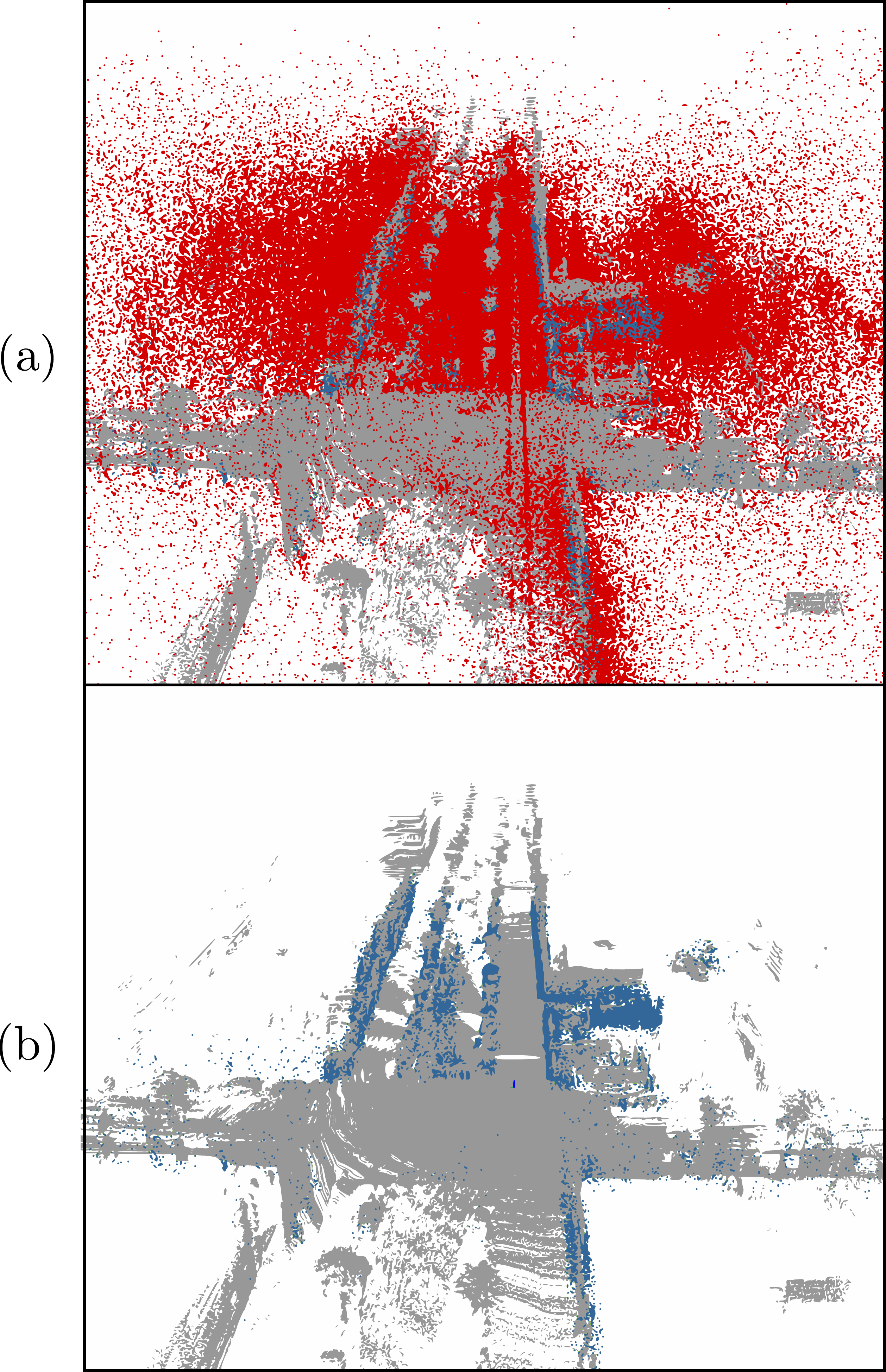}
	\caption{Sequence 00 with (a) radar raw detections (red), LiDAR (grey) and (b) corrected labels (${y} (\mathbf{p}_{i,t}) = 1 $) in blue.}
	\label{fig:seq00}
\end{figure}

\textbf{Sequence 00}
Urban crossing scene with buildings, parked cars and vegetation in form of singular trees along the road; see Figure~\ref{fig:seq00}.

\begin{figure}[h!]
	\centering
	\includegraphics[width=0.83\columnwidth]{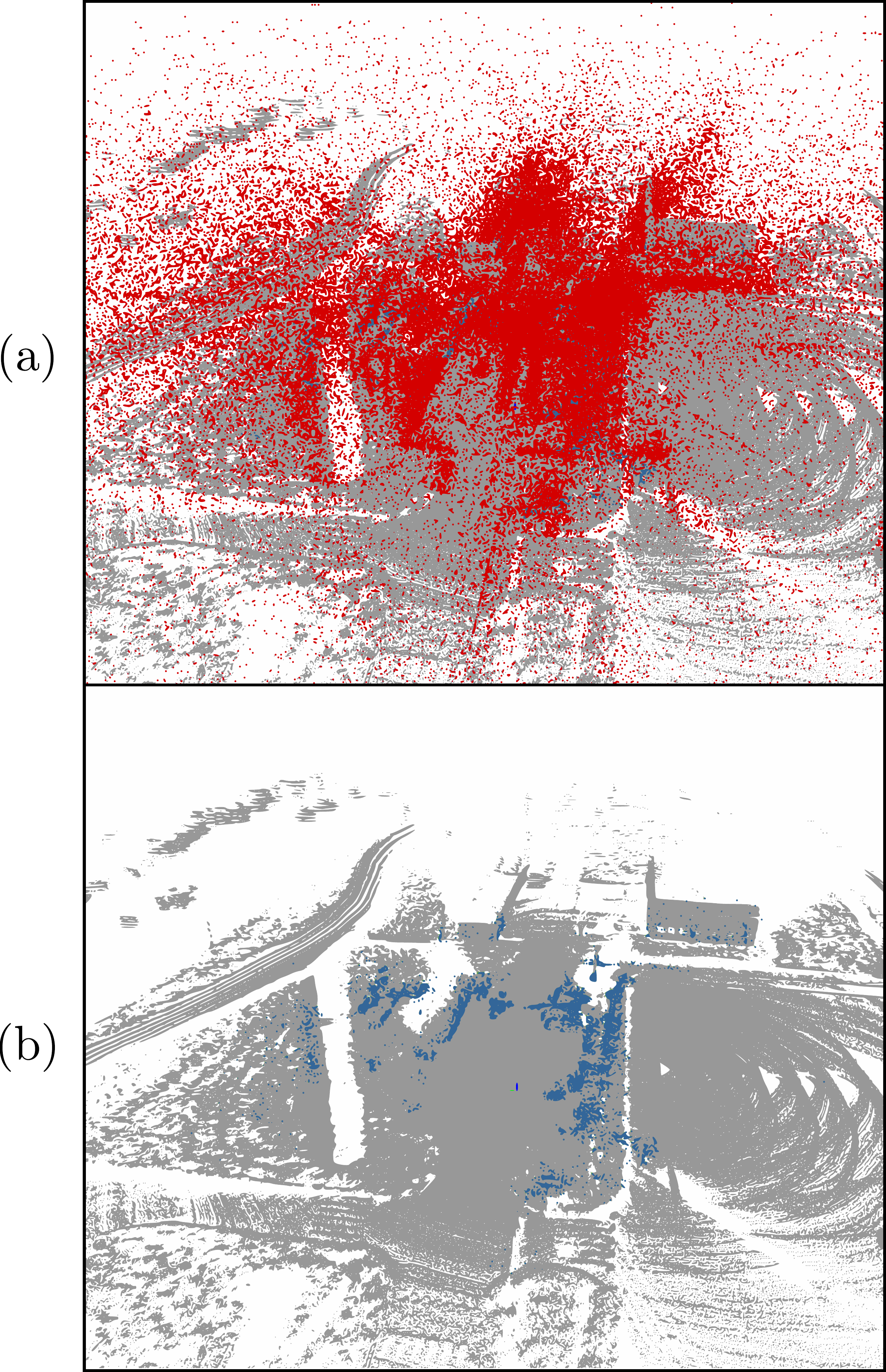}
	\caption{Sequence 01; figure description is equal to Figure~\ref{fig:seq00}.}
	\label{fig:seq01}
\end{figure}

\textbf{Sequence 01}
Scene on open space along parked vehicles. Green area beside street and buildings in background; see Figure~\ref{fig:seq01}.

\begin{figure}[h!]
	\centering
	\includegraphics[width=0.83\columnwidth]{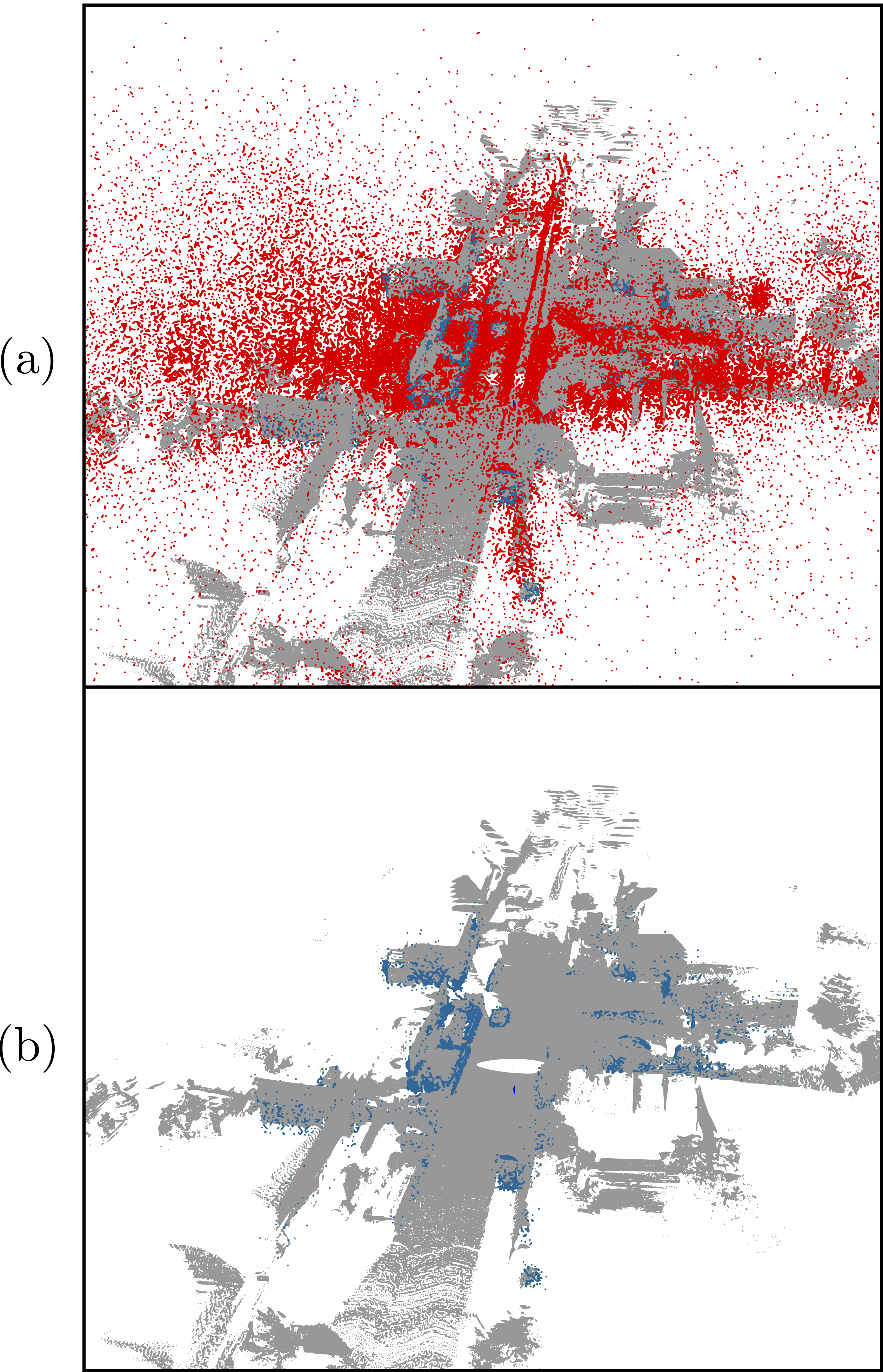}
	\caption{Sequence 02; figure description is equal to Figure~\ref{fig:seq00}.}
	\label{fig:seq02}
\end{figure}

\textbf{Sequence 02}
Straight urban scene, road framed by buildings; Figure~\ref{fig:seq02}.

\begin{figure}[h!]
	\centering
	\includegraphics[width=0.83\columnwidth]{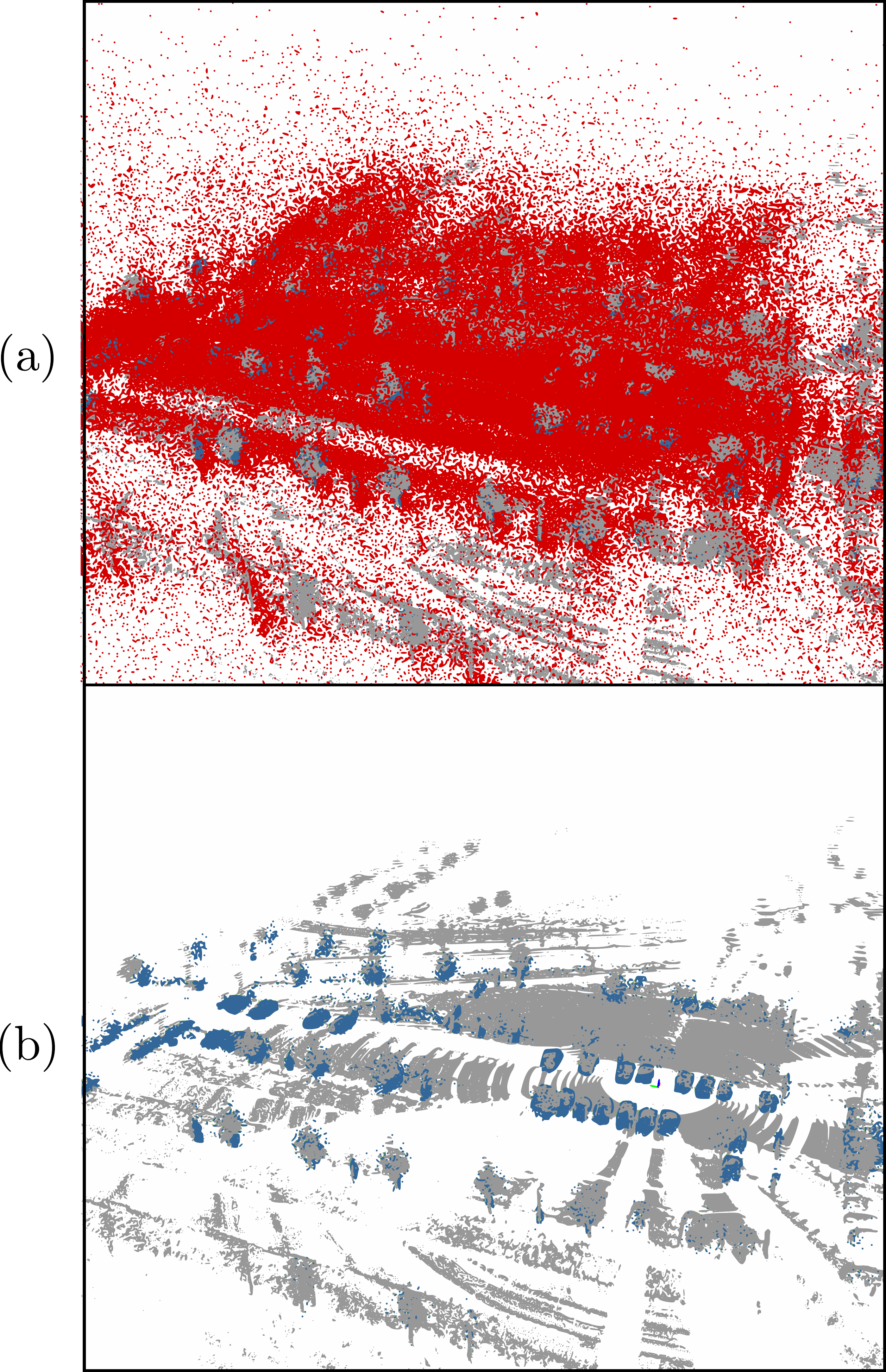}
	\caption{Sequence 03; figure description is equal to Figure~\ref{fig:seq00}.}
	\label{fig:seq03}
\end{figure}

\textbf{Sequence 03}
Public parking lot with parking rows framed by vegetation (bushes, hedges and trees); see Figure~\ref{fig:seq03}.

\begin{figure}[h!]
	\centering
	\includegraphics[width=0.83\columnwidth]{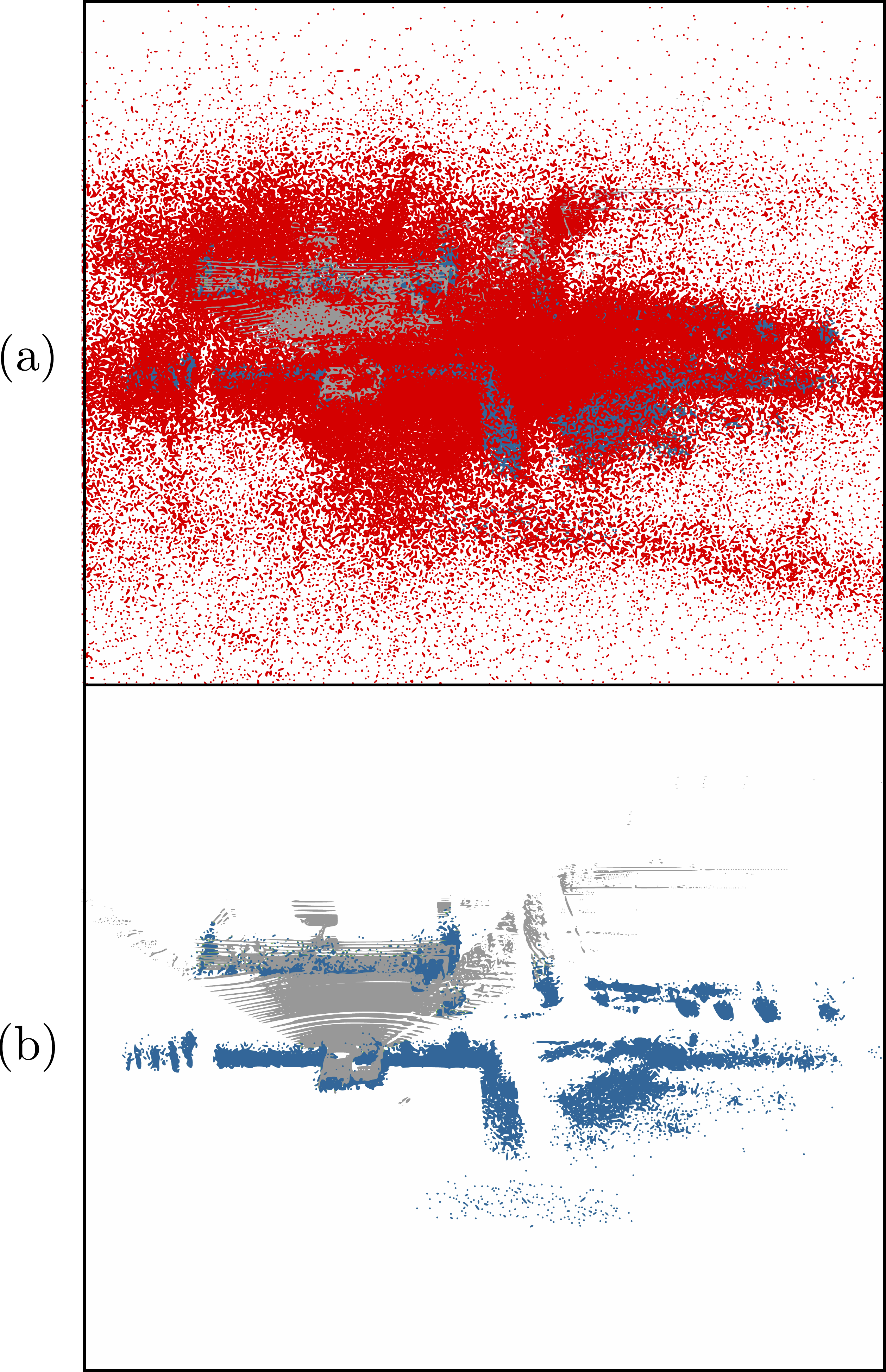}
	\caption{Sequence 04; figure description is equal to Figure~\ref{fig:seq00}.}
	\label{fig:seq04}
\end{figure}

\textbf{Sequence 04}
Exit of a garage and maneuver in front of building; see Figure~\ref{fig:seq04}.

\begin{figure}[h!]
	\centering
	\includegraphics[width=0.83\columnwidth]{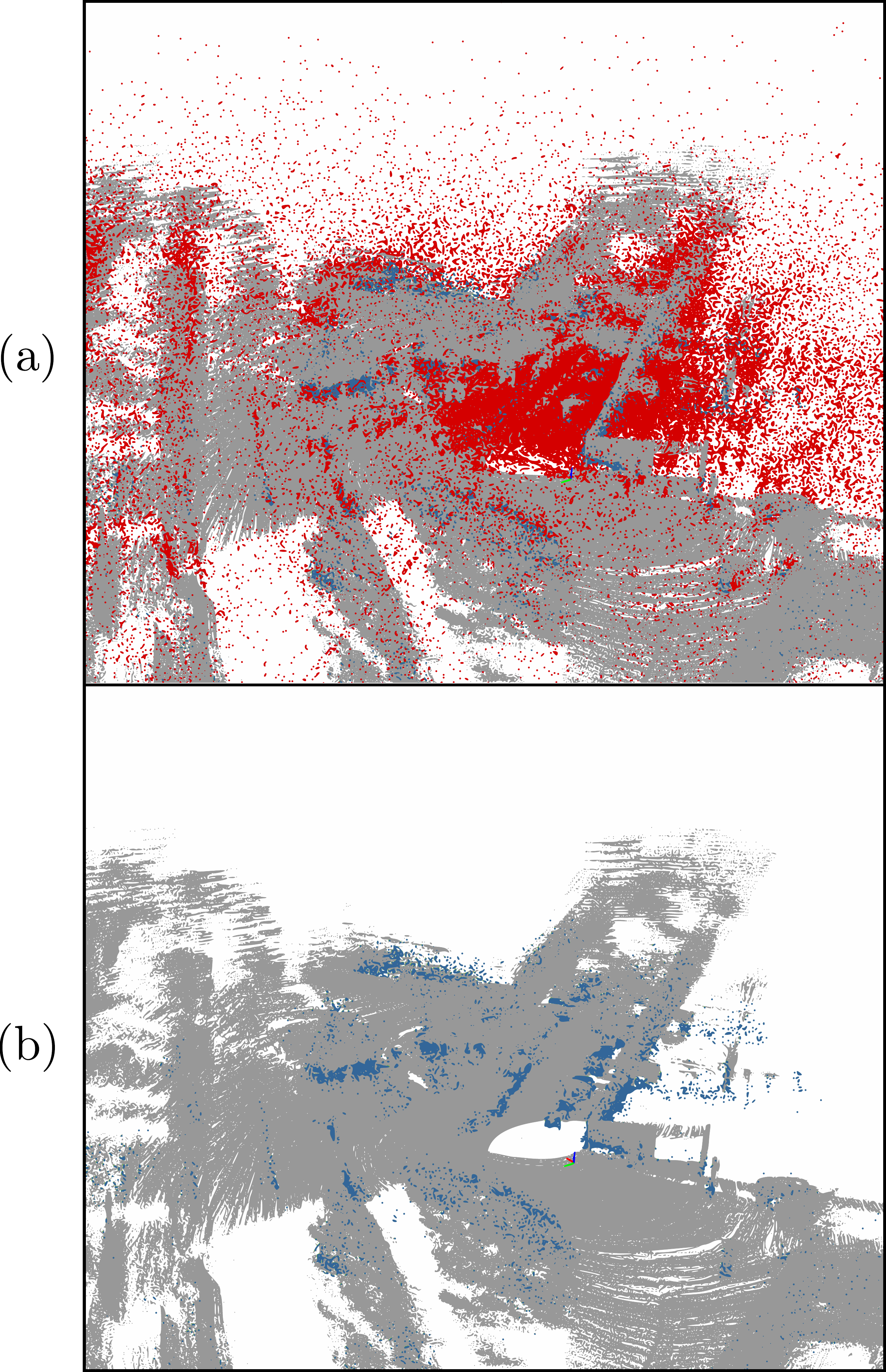}
	\caption{Sequence 05; figure description is equal to Figure~\ref{fig:seq00}.}
	\label{fig:seq05}
\end{figure}

\textbf{Sequence 05}
Urban crossing scene with open space around crossing, road framed by buildings; see Figure~\ref{fig:seq05}. 

\begin{figure}[h!]
	\centering
	\includegraphics[width=0.83\columnwidth]{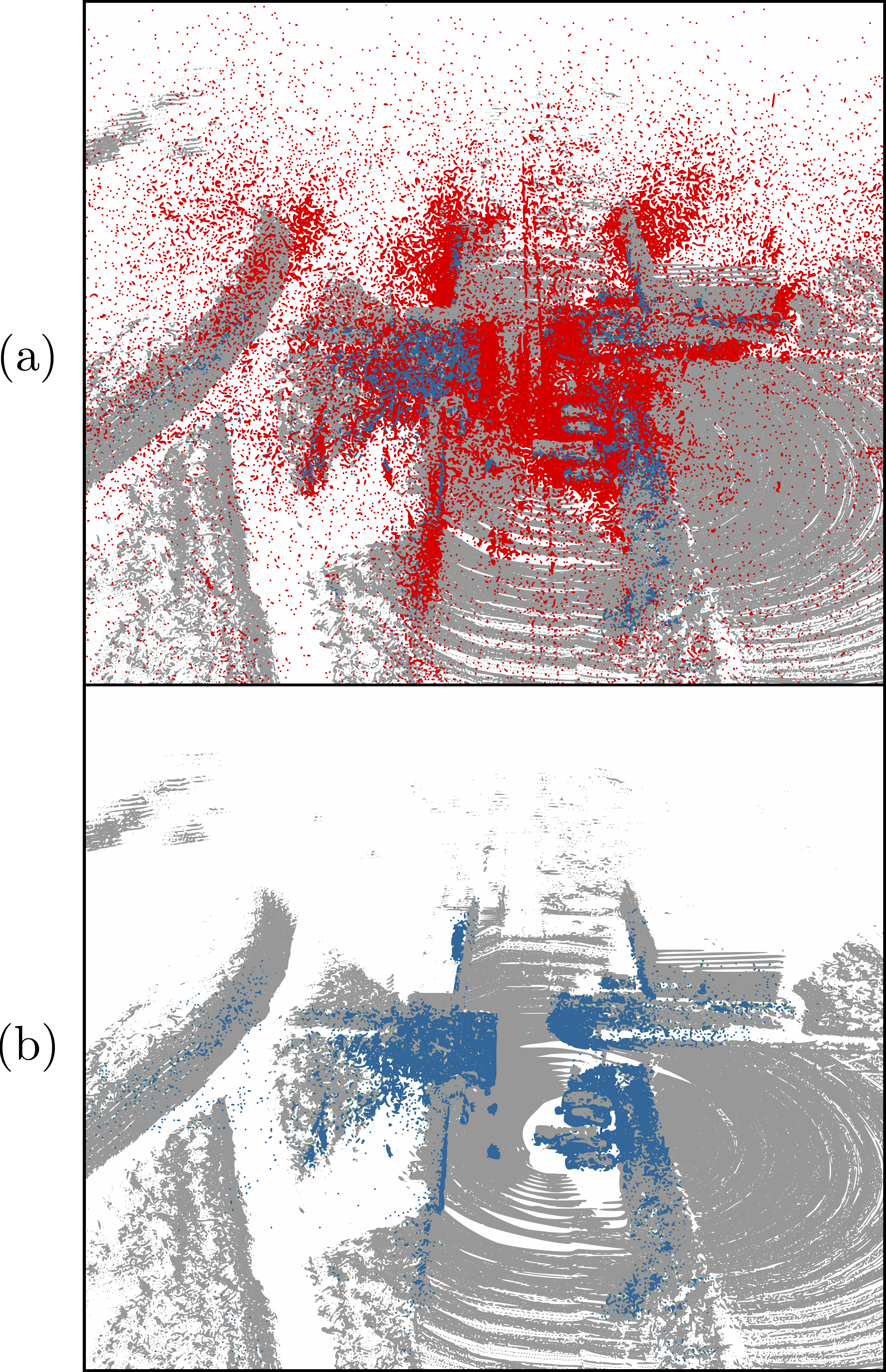}
	\caption{Sequence 06; figure description is equal to Figure~\ref{fig:seq00}.}
	\label{fig:seq06}
\end{figure}

\textbf{Sequence 06}
Scene on open space along parked vehicles. Green area beside street and buildings in background; see Figure~\ref{fig:seq06}. Other driving direction as in Sequence 01. Overlapping area with sequence 04.

\begin{figure}[h!]
	\centering
	\includegraphics[width=0.83\columnwidth]{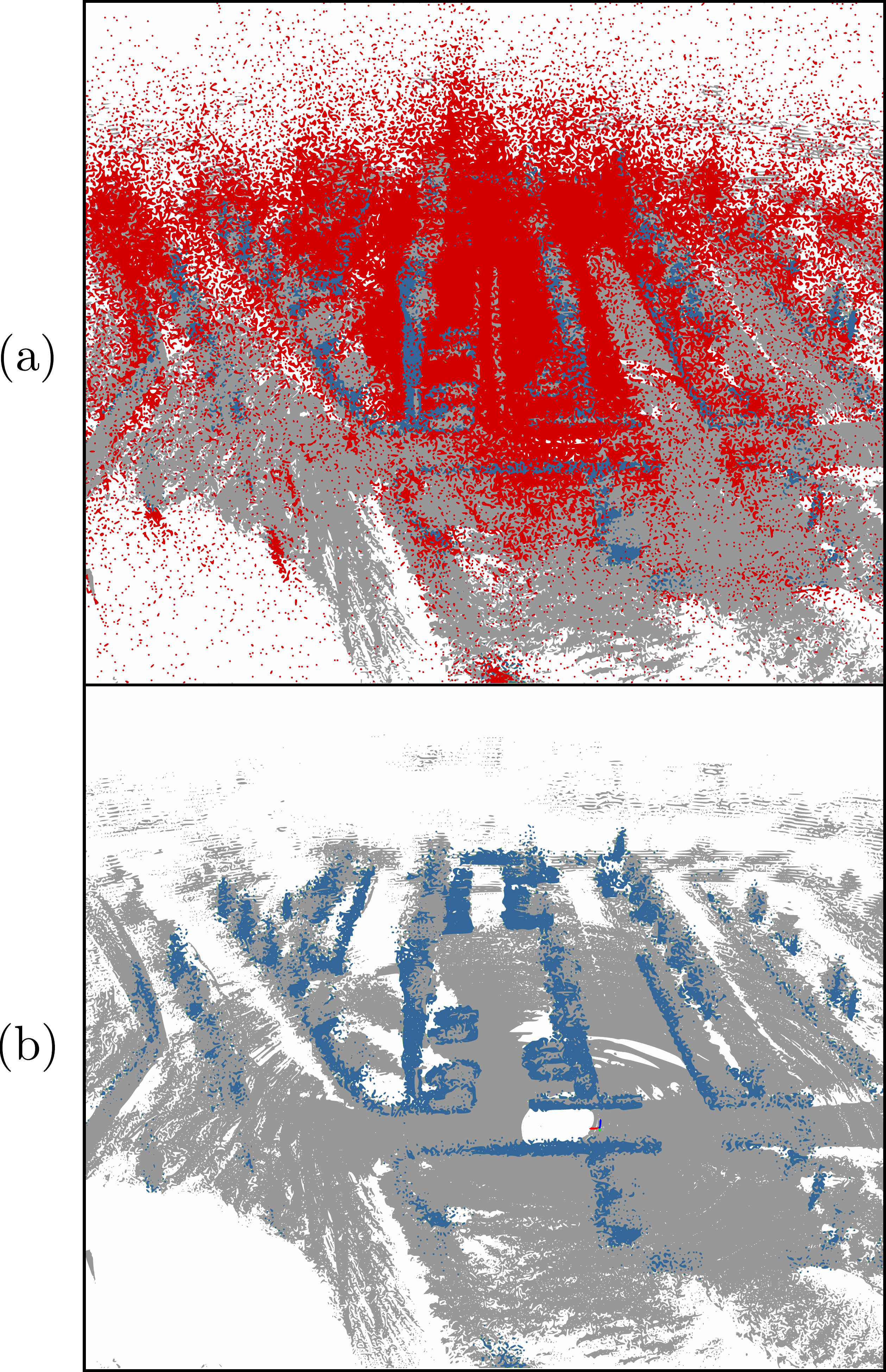}
	\caption{Sequence 07; figure description is equal to Figure~\ref{fig:seq00}.}
	\label{fig:seq07}
\end{figure}

\textbf{Sequence 07}
Public parking lot with parking rows framed by vegetation (bushes, hedges, and trees); see Figure~\ref{fig:seq07}.

\begin{figure}[h!]
	\centering
	\includegraphics[width=0.83\columnwidth]{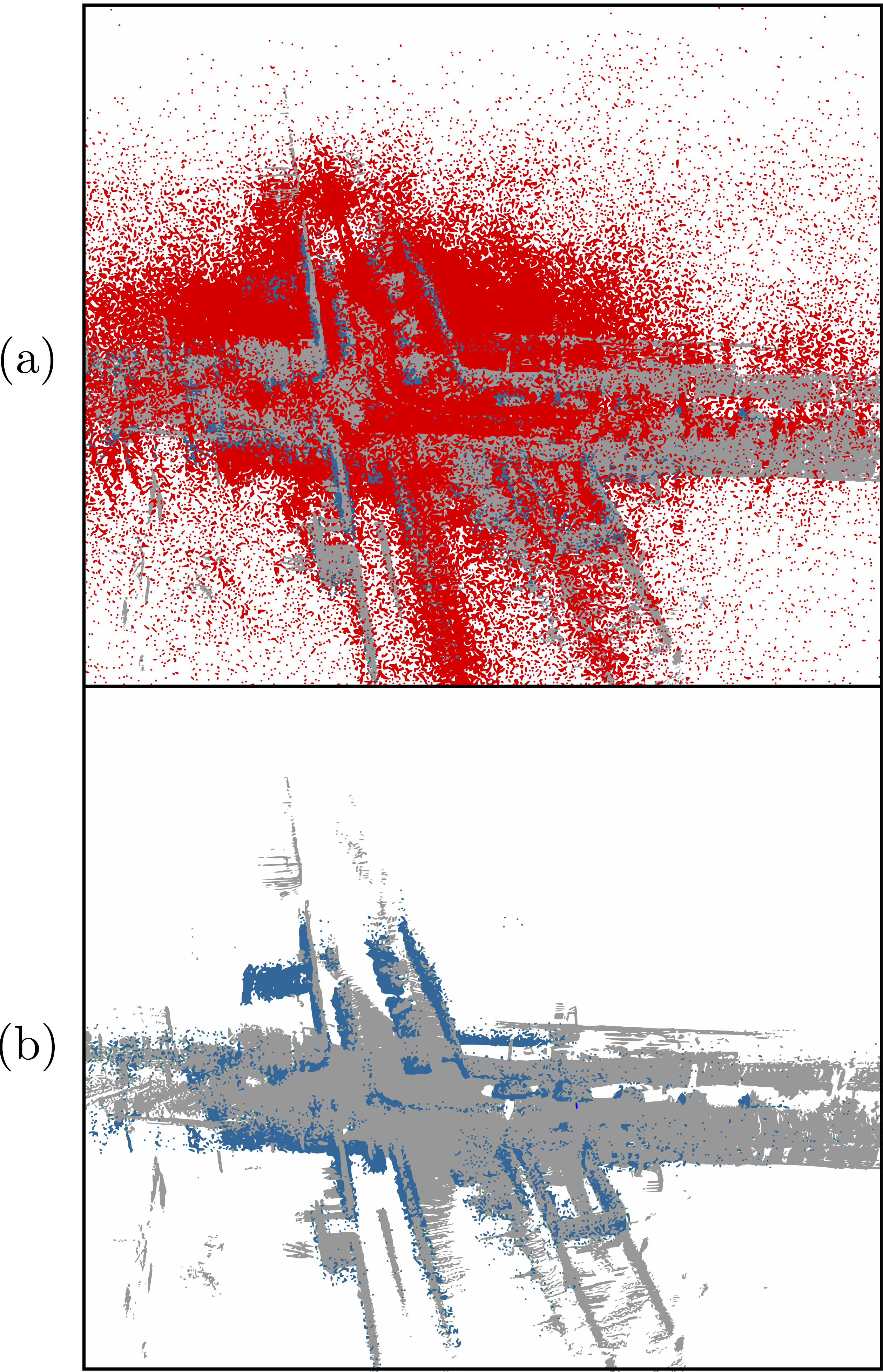}
	\caption{Sequence 08; figure description is equal to Figure~\ref{fig:seq00}.}
	\label{fig:seq08}
\end{figure}

\textbf{Sequence 08}
Urban crossing scene with buildings, parked cars and vegetation in form of singular trees in crossbreeding road; see Figure~\ref{fig:seq08}.

\begin{figure}[h!]
	\centering
	\includegraphics[width=0.83\columnwidth]{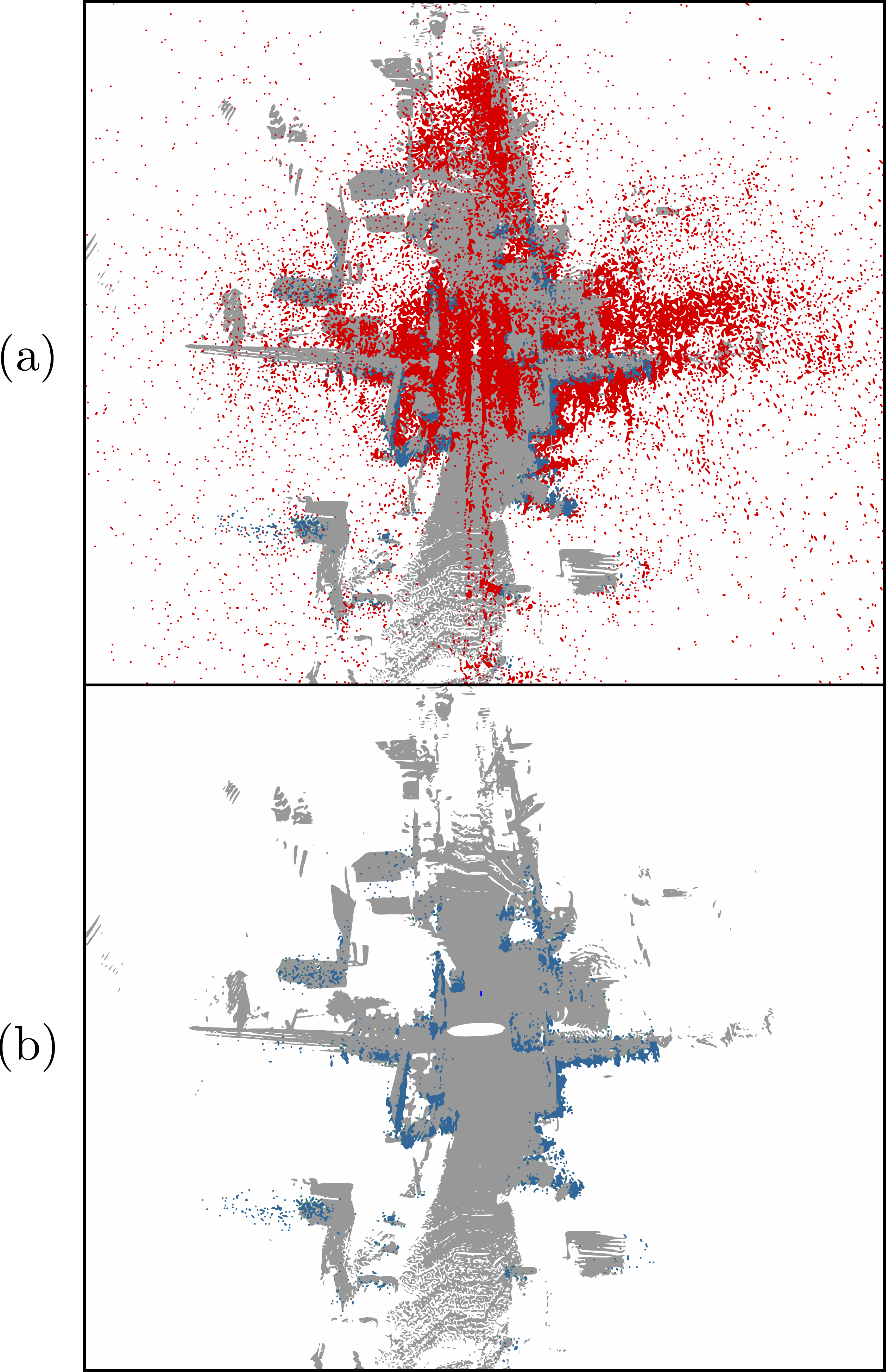}
	\caption{Sequence 09; figure description is equal to Figure~\ref{fig:seq00}.}
	\label{fig:seq09}
\end{figure}

\textbf{Sequence 09}
Residential area with single-family houses and front yards as road frame; see Figure~\ref{fig:seq09}.

\begin{figure}[h!]
	\centering
	\includegraphics[width=0.83\columnwidth]{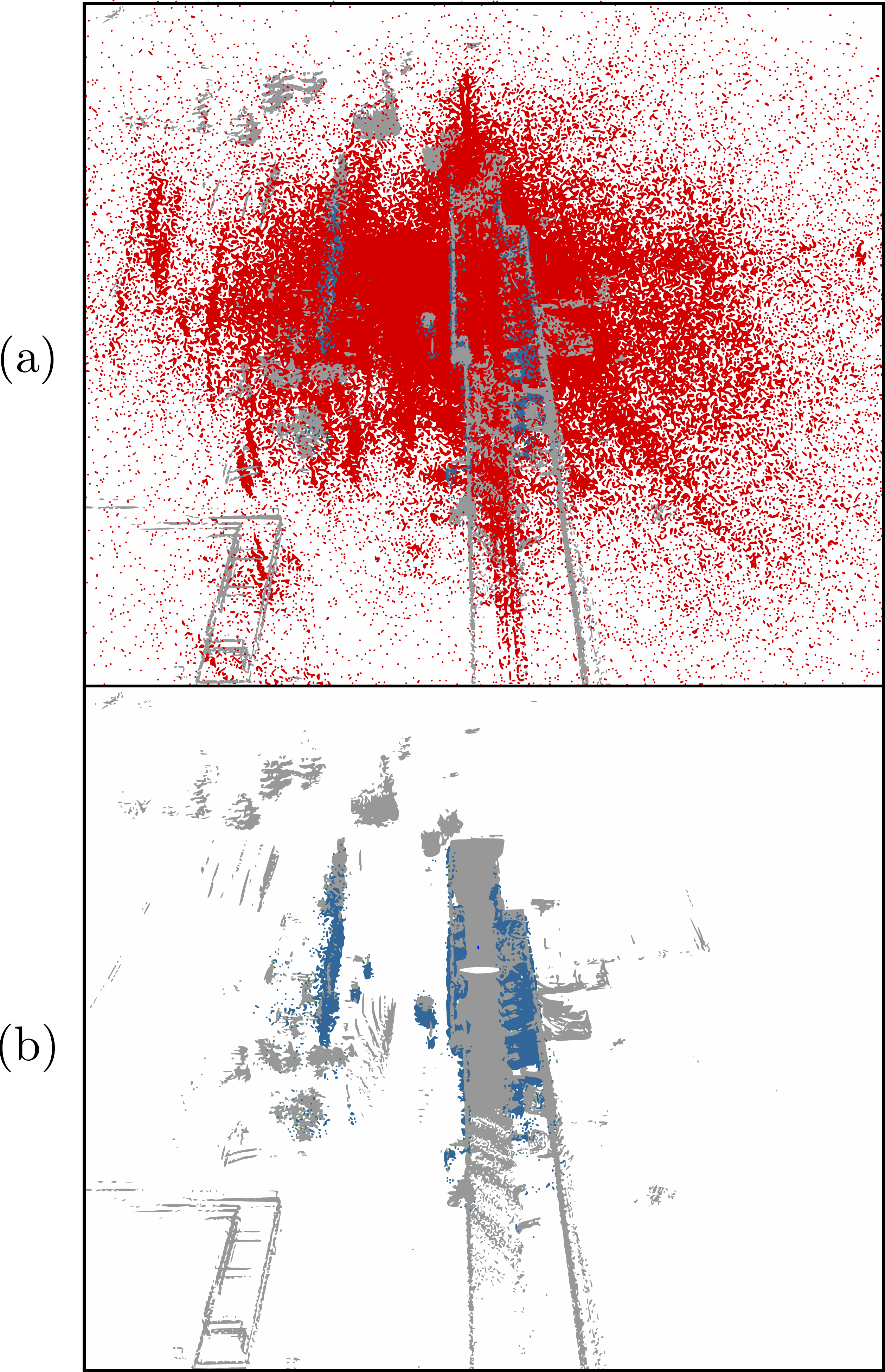}
	\caption{Sequence 10; figure description is equal to Figure~\ref{fig:seq00}.}
	\label{fig:seq10}
\end{figure}

\textbf{Sequence 10}
Urban area, straight drive along row of fishbone shaped cars on one side, opposed to a fence; see Figure~\ref{fig:seq10}. The fence was labeled plausible in order to represent a impassable wall.


\end{document}